\documentclass[lettersize,journal]{IEEEtran}
\usepackage{amsmath,amsfonts}
\usepackage{algorithmic}
\usepackage{algorithm}
\usepackage{booktabs} 
\usepackage{array}
\usepackage[caption=false,font=normalsize,labelfont=rm,textfont=rm]{subfig}
\usepackage{textcomp}
\usepackage{stfloats}

\usepackage{url}
\usepackage{verbatim}
\usepackage{graphicx}
\usepackage{epstopdf}
\usepackage{cite}
\usepackage{adjustbox}
\usepackage{float}
\usepackage{multirow}
\usepackage{pifont}
\usepackage{amssymb}
\usepackage[hidelinks]{hyperref}
\usepackage[table,xcdraw]{xcolor}
\begin{document}
\setlength{\textfloatsep}{5pt}

\title{EarthGPT: A Universal Multi-modal Large Language Model for Multi-sensor Image Comprehension in Remote Sensing Domain}

\author{
    Wei Zhang\textsuperscript{*}, Miaoxin Cai\textsuperscript{*},~\IEEEmembership{Student Member,~IEEE,} Tong Zhang,~\IEEEmembership{Student Member,~IEEE,} \\Yin Zhuang,~\IEEEmembership{Member,~IEEE,} and Xuerui Mao\textsuperscript{\dag}
    \thanks{ * The authors contributed equally to this work.}
    \thanks{ {\dag} Corresponding author: Xuerui Mao.}
    \thanks{Wei Zhang is with the Advanced Research Institute of Multidisciplinary Sciences, Beijing Institute of Technology, Beijing 100081, China, and also with the School of Mechatronical Engineering, Beijing Institute of Technology, Beijing 100081, China. (e-mail: w.w.zhanger@gmail.com).} 
    \thanks{Xuerui Mao is with the Advanced Research Institute of Multidisciplinary Sciences, Beijing Institute of Technology, Beijing 100081, China, and with the School of Mechatronical Engineering, Beijing Institute of Technology, Beijing 100081, China, and also with Yangtze Delta Region Academy of Beijing Institute of Technology, Jiaxing 314003, China. (e-mail: xmao@bit.edu.cn).
    }
    \thanks{Miaoxin Cai, Tong Zhang, and Yin Zhuang are with the National Key Laboratory of Science and Technology on Space-Born Intelligent Information Processing, Beijing Institute of Technology, Beijing 100081, China. (e-mail: 3120220667@bit.edu.cn, bit\_zhangtong@163.com, yzhuang@bit.edu.cn).
    }
}

\maketitle




\begin{abstract}
Multi-modal large language models (MLLMs) have demonstrated remarkable success in vision and visual-language tasks within the natural image domain. Owing to the significant diversities between the natural and remote sensing (RS) images, the development of MLLMs in the RS domain is still in the infant stage. To fill the gap, a pioneer MLLM named EarthGPT integrating various multi-sensor RS interpretation tasks uniformly is proposed in this paper for universal RS image comprehension. Firstly, a visual-enhanced perception mechanism is constructed to refine and incorporate coarse-scale semantic perception information and fine-scale detailed perception information. Secondly, a cross-modal mutual comprehension approach is proposed, aiming at enhancing the interplay between visual perception and language comprehension and deepening the comprehension of both visual and language content. Finally, a unified instruction tuning method for multi-sensor multi-task in the RS domain is proposed to unify a wide range of tasks including scene classification, image captioning, region-level captioning, visual question answering (VQA), visual grounding, object detection, etc. More importantly, a dataset named MMRS-1M featuring large-scale multi-sensor multi-modal RS instruction-following is constructed, comprising over 1M image-text pairs based on 34 existing diverse RS datasets and including multi-sensor images such as optical, synthetic aperture radar (SAR), and infrared. The MMRS-1M dataset addresses the drawback of MLLMs on RS expert knowledge and stimulates the development of MLLMs in the RS domain. Extensive experiments are conducted, demonstrating the EarthGPT's superior performance in various RS visual interpretation tasks compared with the other specialist models and MLLMs, proving the effectiveness of the proposed EarthGPT and offering a versatile paradigm for open-set reasoning tasks. Our code and dataset are available at \href{https://github.com/wivizhang/EarthGPT}{\textit{https://github.com/wivizhang/EarthGPT}}.

\end{abstract}

\begin{IEEEkeywords}
Multi-modal large language model (MLLM), remote sensing (RS), instruction-following, multi-sensor.
\end{IEEEkeywords}

\begin{figure*}
	\centering
		\includegraphics[scale=0.270]{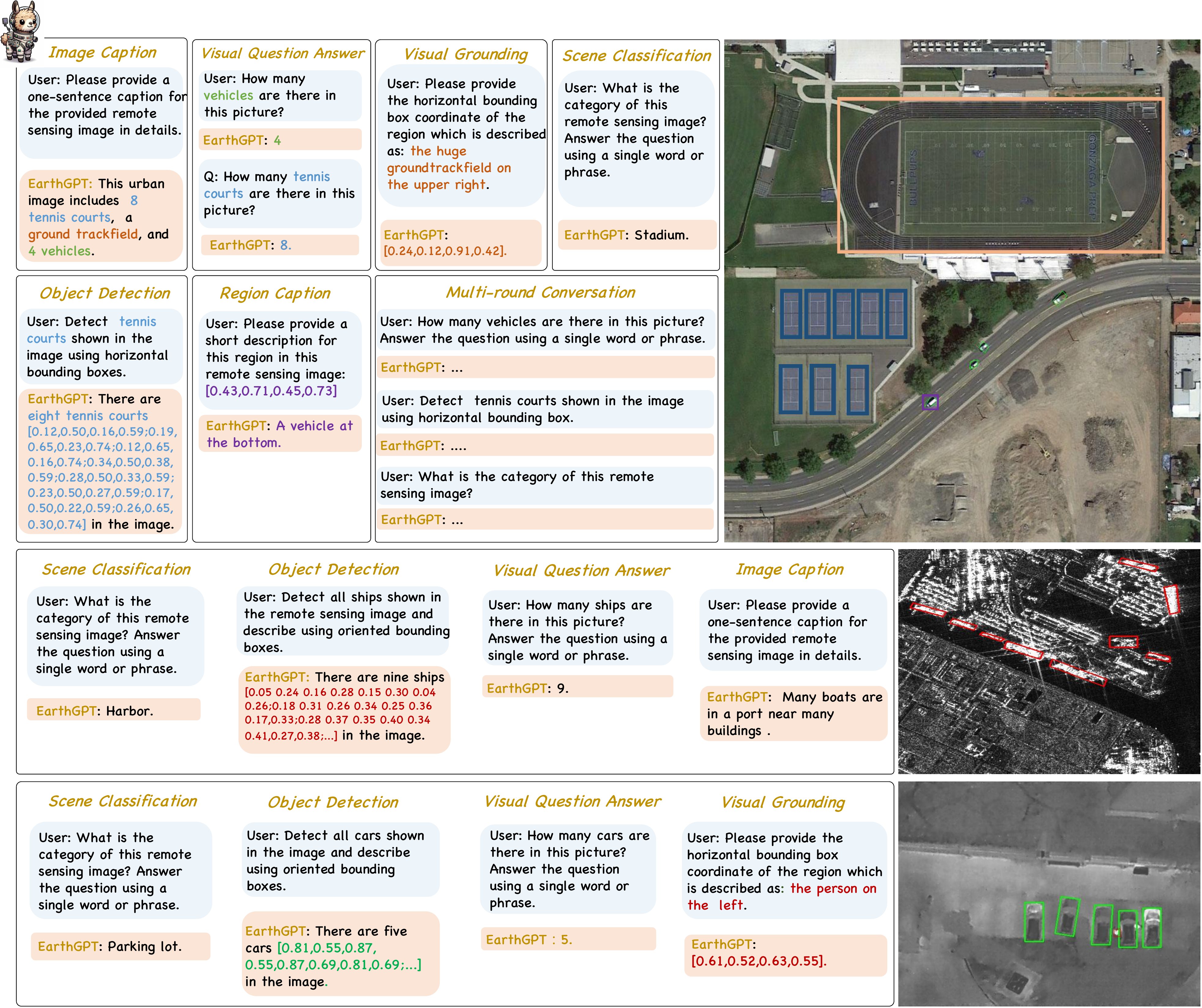}
	\caption{EarthGPT is a pioneering model designed to seamlessly unify multi-sensor and diverse RS intelligent interpretation tasks in a unified framework, guided by user language instructions, and is versatile at performing visual-language dialogues across optical, SAR, and infrared images. EarthGPT's capabilities extend to a wide range of tasks including scene classification, image description, visual question answering, target description, visual localization, and object detection.}
	\label{FIG:EarthGPT}
\end{figure*}

\vspace{1cm}
\section{Introduction}
\IEEEPARstart{U}{nifying} {\normalsize multi-task interpretation in the remote sensing (RS) domain is crucial in practical application as real-world scenarios often demand comprehensive analyses to make informed decisions. Although deep learning methods in RS have been successful in RS visual analysis tasks\cite{ma2019deep,zhu2017deep,zhang2022consecutive}, current methods mainly follow a one-task-one-architecture paradigm which limits their ability to handle multi-sensor RS images, multiple tasks, and generalize to open-set reasoning. 

Most recently, the development of multi-modal large language models (MLLMs)~\cite{chen2022visualgpt,li2023blip,liu2023visual,alayrac2022flamingo}, has shown great success in the natural image domain, as they not only possess robust language interaction capabilities but also exhibit impressive multi-task reasoning skills in real-world scenarios~\cite{zhao2023mmicl}, compared to smaller and domain-specific models tailored for particular tasks. These MLLMs showed effectiveness in tasks like detailed image description, question answering, spatial localization guided by language instructions, etc. The remarkable multi-modal reasoning ability allows MLLMs to generalize well in new situations and demonstrate powerful zero-shot capabilities across open-set tasks~\cite{zhu2023minigpt}.

However, owing to the significant diversities between the natural and multi-sensor RS images such as the imaging conditions, environments, scales, and object viewpoints, there are significant challenges for the application of MLLMs in the RS domain. Although rare MLLM works such as RSGPT\cite{hu2023rsgpt} and GeoChat\cite{kuckreja2023geochat} have been proposed to explore the MLLMs for RS and to solve multiple tasks, they still have limitations. Notably, RSGPT has to tune the model for each task individually, leading to the lack of generalizable ability in the open-set domain. RSGPT also struggles with tasks like classification, detection, and visual grounding. In addition, GeoChat is trained on optical RS images and lacks the multi-sensor datasets to realize synthetic aperture radar (SAR), and infrared modalities comprehension. It is clear that the study of MLLM in the RS domain is still in its infancy. To address these problems thoroughly, this paper aims to unify a wide range of RS tasks and concentrates on constructing a large-scale multi-modal dataset containing multi-sensor RS images based on diverse existing RS datasets.

To leverage the robust generalization and emergent ability of large language models (LLMs), a versatile MLLM called EarthGPT is proposed for multi-sensor RS image comprehension, unifying various RS interpretation tasks effectively. As illustrated in Fig. \ref{FIG:EarthGPT}, EarthGPT can serve as an intelligent assistant capable of efficiently handling a wide range of RS tasks through language interaction. In EarthGPT, three key techniques are proposed. Firstly, a visual-enhanced perception mechanism is constructed to mix various visual backbones. Specifically, the mixed backbones refine coarse-scale semantic perception information and fine-scale detailed perception information, thereby enhancing visual comprehension. Secondly, a cross-modal mutual comprehension approach is proposed. By directly concatenating the visual features with language features to generate multi-modal input for LLM, then unfreezing the self-attention and RMSNorm layer of transformer blocks to train on natural common datasets, visual-language alignment is implemented, and the interplay comprehension between visual and language content is deepened. Finally, a unified instruction tuning method for multi-sensor and multi-task in the RS domain is proposed, by continuing fine-tuning LLM using the bias tuning strategy, MLLM is endowed with the capability of unifying a wide range of tasks including scene classification, image captioning, region-level captioning, visual question answering (VQA), visual grounding, horizontal bounding box (HBB) object detection, and oriented bounding box (OBB) object detection. More importantly, a large-scale dataset called MMRS-1M which is a multi-sensor multi-modal RS instruction-following dataset, is created. MMRS-1M comprises over 1M image-text pairs transformed from 34 existing diverse RS datasets and includes multi-sensor images such as optical, SAR, and infrared. This dataset is tailored to the unique visual modalities and geographical characteristics of the RS domain, effectively mitigating the lack of RS domain expertise of MLLMs. Furthermore, MMRS-1M serves as a catalyst for the advancement of MLLMs in the RS domain.

Extensive experiments are conducted in multiple RS datasets. It shows that the proposed EarthGPT surpasses the state-of-the-art (SOTA) specialist models and MLLMs in most RS tasks. In particular, for the image captioning, VQA, and visual grounding task in the supervised setting, EarthGPT shows a notable improvement in the NWPU-Captions\cite{cheng2022nwpu}, CRSVQA~\cite{zhang2023multi}, and DIOR-RSVG~\cite{10056343} datasets compared with other specialist models. We also assess generalization ability in the open-set domain for the proposed EarthGPT. For the zero-shot scene classification task, EarthGPT achieves 77.37$\%$ and 74.72$\%$ accuracy on the CLRS~\cite{li2020clrs} and NaSC-TG2~\cite{zhou2021nasc} datasets, respectively, far exceeding other MLLMs. On the MAR20~\cite{yu2022mar20} detection dataset, EarthGPT achieves AP@40$\%$ metrics of 90.47$\%$ accuracy, outperforming other MLLMs and open-set detection models.  In conclusion, experimental results demonstrate EarthGPT’s superior performance in a wide range of RS multi-sensor image comprehension tasks, and robust generalization capability in open-set reasoning tasks.
 
In summary, the contributions of this paper are as follows.
\begin{itemize}
\item{To our best knowledge, the largest multi-modal multi-sensor RS instruction-following dataset named MMRS-1M is constructed, consisting of over 1M image-text pairs that include optical, SAR, and infrared RS images. This dataset effectively mitigates the lack of RS domain expertise in MLLMs.}

\item{A pioneer MLLM called EarthGPT is proposed for multi-sensor RS image comprehension, and is capable in a wide range of visual-language RS tasks uniformly. It consists of three techniques. The first is a visual-enhanced perception mechanism to refine coarse-scale and fine-scale visual perception information. The second is a cross-modal mutual comprehension approach that realizes the mutual comprehension of visual and language content. The last is a unified instruction tuning method designed for multi-sensor RS image comprehension and a wide range of RS visual tasks.}

\item{Extensive experiments demonstrate the EarthGPT's superior performance in multi-sensor RS visual interpretation tasks compared with the other specialist models and MLLMs in scene classification, image captioning, region-level captioning, VQA, visual grounding, and object detection. EarthGPT therefore represents a significant advance of MLLMs in the field of RS and contributes a versatile paradigm for RS visual-language mutual comprehension as well as the open-set reasoning ability.}
\end{itemize}

\section{Related Work}
\subsection{MLLMs}
The continuous emergence of LLMs has fueled rapid advancements in natural language processing, showcasing extraordinary capabilities in language modeling across diverse contexts. Pre-training on massive text corpora, fine-tuning on specific domains, and the continual expansion of model parameters have enabled LLMs to achieve new milestones in various benchmark tests. Notably, OpenAI's ChatGPT~\cite{openai2023gpt4} addresses users' needs across different scenarios through human-machine dialogues and text generations. The LLaMA-guided fine-tuning models~\cite{touvron2023llama_a,zhang2023llama,touvron2023llama_b} have gained favor among LLM researchers. In addition, GPT-4-LLM ~\cite{peng2023instruction} demonstrates significant improvements when provided with high-quality instruction-following datasets. As a fundamental component of general intelligence, multi-modal perception is a crucial step toward achieving universal models. Researchers~\cite{chen2022visualgpt,li2023blip,liu2023visual,alayrac2022flamingo} are currently devoted to incorporating multimodal data beyond text into LLM to support specific tasks across various modalities. Models like VisualGPT~\cite{chen2022visualgpt}, BLIP~\cite{li2023blip}, Flamingo~\cite{alayrac2022flamingo}, and Kosmos-1~\cite{huang2023language} showcase strong multimodal reasoning potential after aligning LLMs with image modalities. To achieve modality alignment, many works, such as MiniGPT-4~\cite{zhu2023minigpt}, LLAMA-Adapter V2~\cite{gao2023llama}, mPLUG-Owl~\cite{ye2023mplug}, employed projection layers, zero-shot attention mechanism, and intermediate networks to fuse LLaMA and visual modality. With the continuous progress of MLLMs, additional modalities such as audio, video~\cite{lyu2023macaw,tang2023codi}, 3D point clouds~\cite{xu2023pointllm,guo2023point,hong20233d} are being integrated and aligned. The extraordinary potential of MLLMs is also evident in fields such as LIDAR~\cite{yang2023lidar} and robotics~\cite{ahn2024autort}. 

Nevertheless, existing MLLMs are conventionally tailored for the integration of visual and textual elements in natural scenes, missing the capacity to capture the distinctive contextual complexities of the RS domain. Those MLLMs's adaptability to the complex characteristics of RS data is limited, hindering their capability to effectively realize downstream RS tasks reasoning. To address this gap and create an open-set assistant for RS, EarthGPT is proposed to seamlessly integrate multiple RS tasks and visual modalities from multi-sensors.

\subsection{MLLMs for Remote Sensing}
Previous RS large models primarily employ self-supervised methods~\cite{sun2022ringmo,cong2022satmae} and have progressed in RS visual tasks. However, these models are pre-trained based on visual modalities, lacking alignment with language and struggling with narrow application scenarios, hindering multi-modal understanding and reasoning. Recently, there has been an emergence of MLLMs for RS. For instance, Remoteclip~\cite{liu2023remoteclip} uses contrastive learning pre-training with image-text pairs from existing datasets, demonstrating strong zero-shot classification and image-text retrieval abilities. However, Remoteclip faces challenges in tasks like image captioning, region-level vision grounding, and visual language response due to constraints in the training pattern. RSGPT~\cite{hu2023rsgpt} fine-tunes Instruct-BLIP~\cite{instructblip} on a high-quality image-text pair dataset, showing good image-text caption and VQA capabilities but struggling with tasks like classification, detection, and visual grounding. GeoChat\cite{kuckreja2023geochat} has been proposed to explore  RS MLLMs and to solve multiple tasks. Nevertheless, GeoChat and the above models are trained on optical RS images and lack the generality for multi-sensor visual modalities like SAR, infrared, etc. To achieve more universal multi-modal reasoning in the RS domain and address the limitations of existing multi-modal RS models~\cite{lu2017exploring,zhang2023multi,zheng2021mutual,lobry2020rsvqa,rahnemoonfar2021floodnet,cheng2022nwpu} in open-set dialogue, unifying multiple RS tasks, and multi-sensor image comprehension, this paper focuses on a unified MLLM. The proposed EarthGPT develops MLLM from the natural domain to the RS domain through instruction tuning. Furthermore, to significantly enhance visual-language alignment and fully adapt to the unique characteristics of the RS domain, a visual-enhanced perception mechanism, a cross-modal mutual comprehension mechanism and a unified instruction tuning method for multi-sensor RS image comprehension have been incorporated into EarthGPT. Simultaneously, the constructed MMRS-1M dataset contains various visual modalities.

\subsection{Remote Sensing Datasets}
RS datasets are the core and foundation of RS intelligent interpretation models. Currently, RS datasets mainly focus on classification, detection, segmentation, image captioning, and VQA tasks. For classification, commonly used datasets include AID~\cite{xia2017aid}, EuroSAT~\cite{helber2019eurosat}, NWPU-RESISC45~\cite{7891544}, UCMerced LandUse~\cite{yang2010bag}, and WHU-RS19~\cite{dai2010satellite}. For detection, optical datasets like DIOR~\cite{han2014object}, DOTA~\cite{9560031}, FAIR1M~\cite{sun2022fair1m}, HRRSD~\cite{zhang2019hierarchical}, NWPUVHR10~\cite{li2017rotation}, and SAR datasets like AIR-SARShip-2.0~\cite{xian2019air}, HRSID~\cite{wei2020hrsid}, and SSDD~\cite{zhang2021sar} are used to train expert models. Segmentation datasets typically include Vaihingen, Potsdam, iSAID~\cite{waqas2019isaid}, LoveDA~\cite{NEURIPS-DATASETS-AND-BENCHMARKS2021_4e732ced}, etc. Image captioning datasets include Sydney-Captions~\cite{qu2016deep}, RSICD~\cite{lu2017exploring}, NWPU-Captions~\cite{cheng2022nwpu}, RSITMD~\cite{yuan2022exploring}, and UCM-Captions~\cite{qu2016deep} are used for generating image descriptions or image-text retrieval. VQA datasets mainly consist of Floodnet~\cite{rahnemoonfar2021floodnet}, RSVQA-LR~\cite{lobry2020rsvqa}, RSVQA-HR~\cite{lobry2020rsvqa}, RSIVQA~\cite{zheng2021mutual}, and CRSVQA~\cite{zhang2023multi}. The datasets mentioned above primarily emphasize a single visual modality within an individual task, leading to models with limited generalization capabilities and restricted to specific tasks. MLLMs represent a potential solution to tackle the challenge. However, in the current landscape of the RS domain, there is a scarcity of high-quality image-text instruction-following data. This scarcity poses a significant obstacle when it comes to training multi-modal intelligent dialogue assistants that can achieve a high level of alignment between visual and language, unify various tasks into a framework, and integrate different multi-sensor visual modalities effectively. 

To address this challenge, we have constructed a dataset called MMRS-1M containing more than 1M image-text pairs, covering tasks such as classification, detection, image captioning, VQA, visual grounding, etc. MMRS-1M encompasses three visual modalities such as optical, infrared, and SAR. MMRS-1M is designed to promote the continuous development of MLLMs in the RS field.

\begin{figure*}[htbp]
\centering
\includegraphics[width=7in]{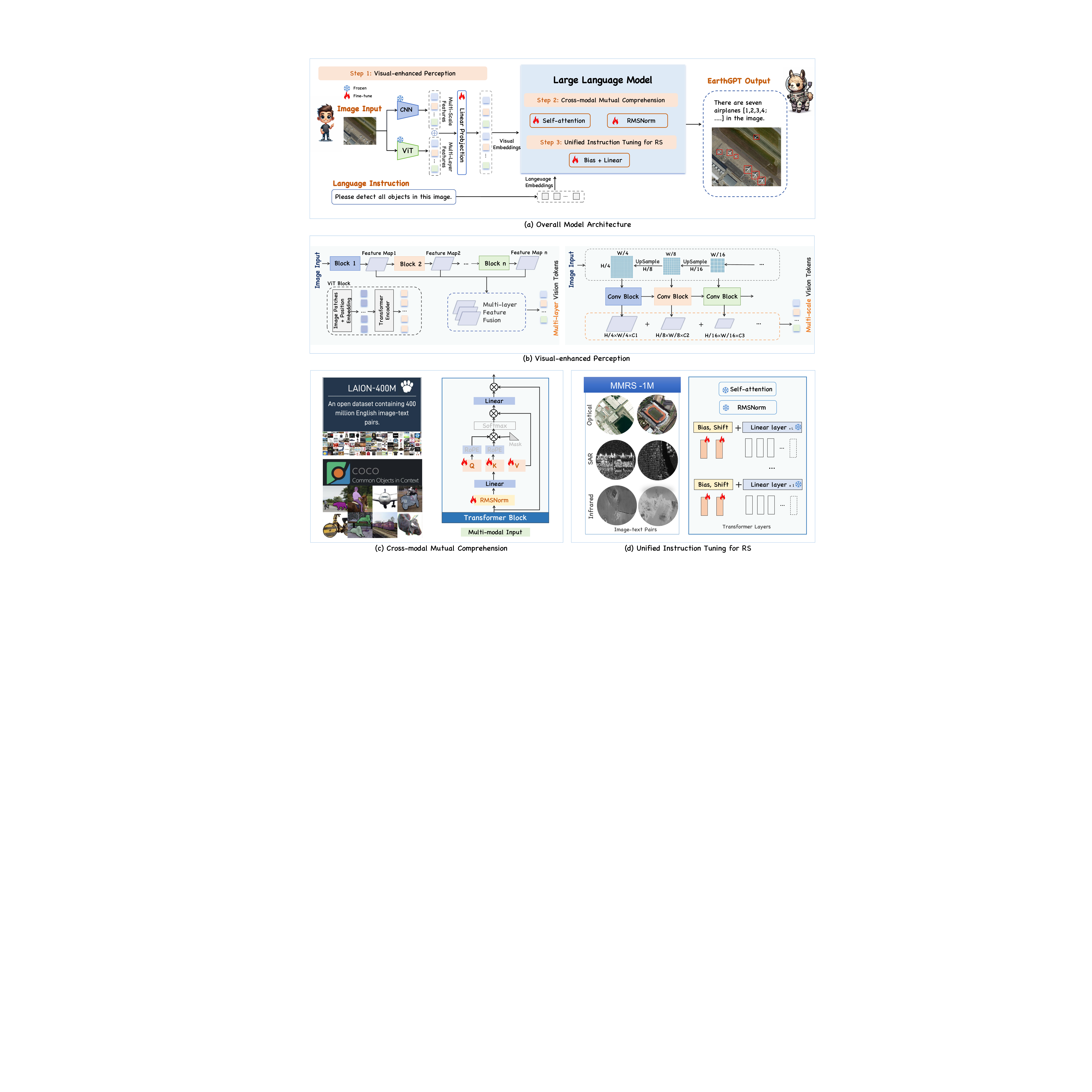}
\caption{(a) Overall model architecture of EarthGPT. (b) Illustration of the visual-enhanced perception mechanism. (c) Illustration of the cross-modal mutual comprehension approach. (d) Illustration of the unified instruction tuning method for RS.}

\label{fig: model}
\end{figure*}

\section{Methodology}
This section introduces the proposed EarthGPT framework. We first overview the overall model architecture (Section III-A). Then, three fundamental techniques in EarthGPT are elaborated involving the visual-enhanced perception mechanism (Section III-B), the cross-modal mutual comprehension approach (Section III-C), and the unified instruction tuning method for RS (Section III-D).

\subsection{Overview}
Our goal is to break down the differences between natural images and RS images that lead to the obstacles for the application of MLLMs in the RS domain and propose a universal MLLM specifically customized for RS. To this end, a unified MLLM called EarthGPT is proposed for RS imagery comprehension. The overall framework of EarthGPT is shown in Fig. \ref{fig: model} (a). Note that three pivotal techniques are developed. Firstly, to tackle the challenge that RS images commonly contain various disturbances that compromise clarity, the visual-enhanced perception mechanism is proposed, as shown in step 1. Specifically, ViT and CNN backbones are designed to refine global and detailed visual features through multi-layer and multi-scale visual perception. Then, twofold visual tokens are concatenated and projected to ensure dimension alignment with language instruction tokens. The hybrid visual tokens contain both neighboring dependencies and long-range visual interactions, offering subtle local insights and extensive regional correlations. Subsequently, the cross-modal mutual comprehension approach is given in step 2. This part is to equip the LLM with fundamental visual comprehension capabilities and achieve visual-language alignment. Particularly, the visual and language tokens are merged as the multi-modal input for the unfrozen LLM tuning to enhance the interaction between the diverse visual and linguistic contexts. Finally, to address the limitations of the existing RS models are specialist ones, unable to handle multi-source and multi-task in one structure. A unified instruction tuning for RS method is developed in step 3. Concretely, the bias tuning of linear layers in LLM is performed on the newly constructed MMRS dataset, achieving the universal visual interpretation in the RS domain. We will introduce each technique as follows in detail.

\subsection{Visual-enhanced Perception}

The core of the visual-enhanced perception mechanism is to utilize the strengths of diverse image encoding models to enhance and refine multi-granularity critical visual information while simultaneously mitigating various disturbances in RS images. The visual-enhanced perception mechanism consists of two modules as shown below. 

\textbf{Multi-Layer Visual Perception.} We adopt the vision transformer (ViT)\cite{dosovitskiy2020image} as the image encoder to extract multi-layer features. In ViT, images are represented as sequences, allowing models to learn image structures independently. First, multi-layer visual intermediate feature maps $\left \{ {V_{a}^{i} } \right \} _{i=1}^n $ are extracted from different encoder layers to capture long-range contextual information, where $n$ represents the number of layers. The original images $I$ are transformed into visual representations $V_{a}^{1}$. Then, all the extracted visual features are concatenated along the channel dimension. Such visual feature fusion from different layers is beneficial to capture subtle differences in RS images and provides a more comprehensive understanding of images. The entire process can be formulated as 
\begin{equation}
V_{a}^{1}  = \mathrm{ViTBlock} _1(I),
\end{equation}
\begin{equation}
V_{a}^{i}  = \mathrm{ViTBlock} _i(V_{a}^{i-1}),~i = 2,...,n,
\end{equation}
\begin{equation}
{{V_a}}  =\mathrm{Concat} \left [V_{a}^{1} ,V_{a}^{2},...V_{a}^{n} \right ].
\end{equation}
In this way, the extracted features both involve spatial-aware information from early layers and semantic indicative from later layers. The multi-layer visual perception is shown at the left of Fig. \ref{fig: model} (b).

\textbf{Multi-scale Visual Perception.} To integrate multi-scale local details into visual representations, the CNN\cite{o2015introduction} backbone is designed as the image encoder. Compared to ViT, CNN excels in extracting localized features, such as edges and textures, through its inherent spatial hierarchies and local receptive fields. The extracted multi-scale visual features are denoted as $\left \{ {V_{b}^{i} } \right \} _{i=1}^m $, where $m$ represents the image scale numbers. As illustrated at the right of Fig. \ref{fig: model} (b), the input of CNN includes multiple spatial resolutions of \(\frac{H}{4} \times \frac{W}{4}\), \(\frac{H}{8} \times \frac{W}{8}\), \(\frac{H}{16} \times \frac{W}{16}\), etc, where \(H \times W\) is the original input image resolution. The convolution blocks convert the inputs into token embeddings \(V_b^1 \in \mathbb{R}^{ \frac{H}{4} \times \frac{W}{4} \times C_1}\), \(V_b^2 \in \mathbb{R}^{ \frac{H}{8} \times \frac{W}{8} \times C_2}\), \(V_b^3 \in \mathbb{R}^{ \frac{H}{16} \times \frac{W}{16} \times C_3}\), etc. Meanwhile, all scale features are transformed into the same channel dimension and concatenated together channel-wisely. The feature extraction process can be written as
\begin{align}
V_{b}^{1}  &= \mathrm{ConvBlock} _1(I_1),
\end{align}
\begin{align}
V_{b}^{i}  &= \mathrm{ConvBlock} _i(V_{b}^{i-1}),~i = 2,...,m,
\end{align}
\begin{align}
{{V_b}} & =\mathrm{Concat} \left [V_{b}^{1} ,V_{b}^{2},...V_{b}^{m} \right ],
\end{align}
where the $I_1$ represents the first scale input. The multi-scale fused visual features encompass information of multi-granularity, enabling the capture of both broad semantics and intricate information. 

After extracting visual features using ViT and CNN, multi-layer and multi-scale features are concatenated channel-wisely. Then, a learnable projection layer is used for dimension alignment with language tokens. The new visual tokens after dimension alignment are denoted as $V_p$. This process is formulated as
 \begin{equation}
V_p =\mathrm{Projection} (\mathrm{Concat} (V_a,V_b)).
\end{equation}
The integration of enhanced visual perception information improves the accuracy and efficiency of the image interpretation.

\subsection{Cross-modal Mutual Comprehension}

The key point of the cross-modal mutual comprehension approach lies in visual and language perception information fusion and delivering the multi-modal input into the LLM for alignment and interaction training.  

Visual tokens are obtained based on the aforementioned visual-enhanced perception mechanism. Meanwhile, following most natural language processing (NLP) models, the language tokenizer is employed to segment language instructions into discrete token embeddings. Subsequently, overall visual embeddings $V_p$ are concatenated with the language instruction embeddings $L_p$ to create LLM input tokens $\mathcal X$ as 
\begin{equation}
 {\mathcal X} =\mathrm{Concat} [\overbrace {V_{p}^{1},V_{p}^{2},~...~,V_{p}^{N_v}}^{{\rm{visual~tokens~\mathit{V_p} }}},\underbrace {L_{p}^{1},L_{p}^{2},~...~,L_{p}^{N_l}}_{\mathrm{language~tokens} ~{L_p}}],
\end{equation}
where $N_v$ represents the token length of visual features, $N_l$ represents the token length of language features, $(V_{p}^{1},V_{p}^{2},~...~,V_{p}^{N_v})$ are the mixed visual tokens from $V_p$,  
 $(L_{p}^{1},L_{p}^{2},~...~,L_{p}^{N_l})$ are the language instruction tokens from $L_p$. 
Then, multi-modal input $\mathcal X$ is fed into LLM for fusion and integration.

In current MLLMs\cite{zhu2023minigpt,chen2022visualgpt,li2023blip,liu2023visual,alayrac2022flamingo}, vision-language understanding is achieved through training on a frozen LLM, designed to avoid expensive full-parameter fine-tuning. However, freezing all the LLM's weights significantly limits its potential for comprehensive cross-modal learning. 
To cope with the challenge of the knowledge gap from different modalities and realize multi-modal mutual comprehension, an unfrozen vision-language alignment strategy is designed. Specifically, the self-attention and normalization layers are unfrozen for training on the most common domain data (LAION-400M\cite{schuhmann2021laion}, COCO Caption\cite{chen2015microsoft}). This strategy aids the LLM in thoroughly understanding multi-modal representations and simultaneously preserves the inherent language response capabilities of LLM through the partly frozen modules. In our method, LLaMA-2~\cite{touvron2023llama} with powerful language understanding capability is adopted as the initial LLM. LLaMA-2 is composed of $L$ Transformer blocks. The Transformer block consists of self-attention, RMSNorm, and FFN layers, whose key module self-attention head is composed of key $\mathbf{K}$, query $\mathbf{Q}$, and the value $\mathbf{V}$, as shown in Fig. \ref{fig: model} (c). The $\mathbf{K}$, $\mathbf{Q}$, and $\mathbf{V}$  are implemented via linear layers. The detailed implementation is as follows
\begin{equation}
\mathbf{Q}(x) = \mathbf{W}_q ~x +{\mathbf{b}_q},
\end{equation}
\begin{equation}
\mathbf{K} (x) = \mathbf{W}_k ~x +{\mathbf{b}_k},
\end{equation}
\begin{equation}
\mathbf{V} (x) = \mathbf{W}_v ~x +{\mathbf{b}_v}.
\end{equation}
The parameters $\mathbf{W}_q$, $\mathbf{W}_k$, $\mathbf{W}_v$, $\mathbf{b}_q$, $\mathbf{b}_k$, and $\mathbf{b}_v$ are updated during the training. The RMSNorm component of the transformer is also unfrozen, and the scale operation $\gamma$ is set as a trainable parameter. The RMSNorm can be expressed as
\begin{equation}
 y = \frac{x}{\sqrt{\text{Mean}(x^2)}+\varepsilon  } *{ \gamma },
 \end{equation}
  where
 \begin{equation}
 \text{Mean}(x^2)=\frac{1}{N} \sum_{i=1}^{N} x_i^2.
  \end{equation}

During the training process, the adopted cross-entropy loss function is described as $\mathcal{L}$, the multi-modal input sequence length is denoted as $N_{vl}$ and the parameters of EarthGPT are represented as $\theta$, where $w_i$ denotes the $i$-th word. The cross-entropy loss function $\mathcal{L}$ can be formulated as
\begin{equation}
\label{deqn_ex1a_loss}
 \mathcal{L} = -\sum_{i=1}^{N_{vl}} \text{log}P(w_i|(w_1,w_2...w_{i-1};\theta ).
\end{equation}

After visual-language alignment, the language-only LLM is converted into an MLLM, which can generate the visual interpretation response based on the integrated multi-modal information. 

\subsection{Unified Instruction Tuning for RS}

After the cross-modal mutual comprehension training, the proposed EarthGPT has basic multi-modal reasoning and dialogue capabilities in the natural domain but struggles to follow instructions and perform inference for RS downstream tasks in the aforementioned stage. To enhance EarthGPT's ability to follow instructions for various downstream tasks and expand its applicability from the natural domain to the RS one, we develop an extensive and unified instruction-following dataset called MMRS-1M. In the MMRS-1M dataset, we have standardized all downstream tasks into the format of VQA instructions. By utilizing MMRS-1M for tuning, the model excels in accurately understanding and executing various instructions in the RS domain. 

Technically, to preserve the visual captioning capability obtained in the previous stage and simultaneously enhance compliance with task instructions, we freeze all the weights from the cross-modal mutual comprehension phase and introduce new learnable parameters into the LLaMA-2 model. Specifically, inspired by LLaMA-Adapter V2 \cite{gao2023llama}, the linear layer has been modified by introducing two learnable parameters including a bias $\beta$ and a shift $\alpha$ into linear layers, as shown in Fig. \ref{fig: model} (d). Given a linear layer $y(x) =\mathbf{W} x$, it can be transformed by incorporating the bias factor and shift factor. The process can be expressed as 
\begin{equation}
\label{deqn_ex1a}
y(x) = \alpha \cdot  ( \mathbf{W}x + \beta ),
\end{equation}
where
\begin{equation}
\label{deqn_ex1b}
\alpha   \sim \mathcal{N}(\mu, \sigma^2),~\beta =\mathrm{Init} (0).
\end{equation}
The bias $\beta$ is initialized with zeros, and the shift $\alpha$ is initialized with a random Gaussian, ensuring training stability and effectiveness. Analogously, the loss function in this stage is given as Eq. (\ref{deqn_ex1a_loss}).

Leveraging the advanced reasoning abilities of LLM and our abundant MMRS-1M dataset, EarthGPT is equipped with versatile skills for multi-sensor visual comprehension in the RS domain guided by the language instructions and shows the potential for real-world practical applications. 

\begin{figure*}[t!]
	\centering
		\includegraphics[scale=0.150]{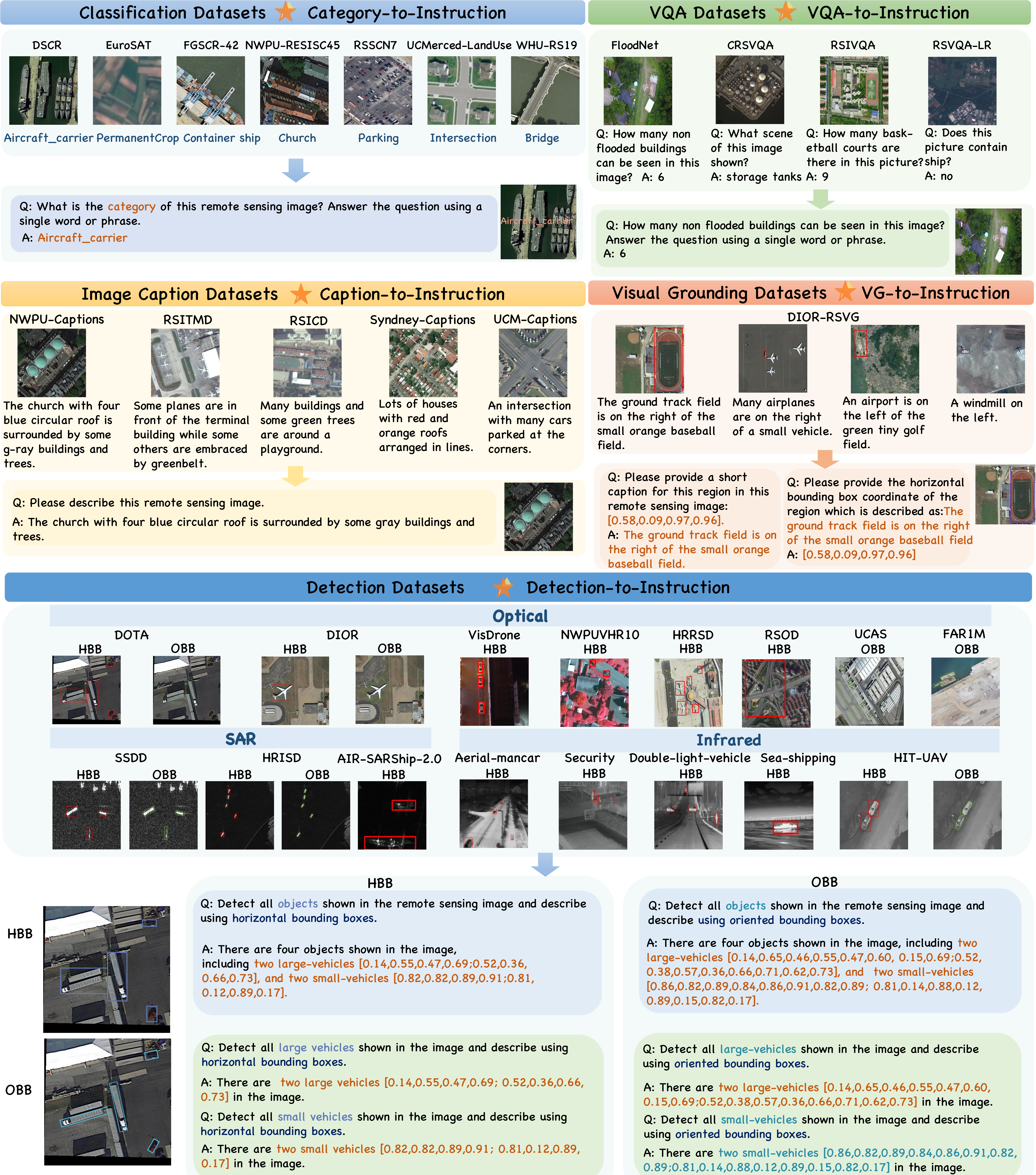}
	\caption{The construction process of MMRS-1M dataset. MMRS-1M contains three visual modalities from multi-sensor (e.g., optical, SAR, and infrared) and five RS vision tasks data(e.g., classification, detection, image caption, VQA, and visual grounding).}
	\label{FIG:datasets}
\end{figure*}

\begin{table}[!b]
\renewcommand{\arraystretch}{1.6}
\caption{Details on the training samples used for the MMRS-1M.}
\label{tab:MMRS-1M Compoents}
\scalebox{0.8}{
\fontsize{10pt}{10pt}\selectfont
\begin{tabular}{cccc}
\hline
Task                                    & Data                & Size & Type   \\ \hline
\multirow{5}{*}{\begin{tabular}[c]{@{}c@{}}Image \\ Captioning\end{tabular}}          & RSICD                      & 24,333& optical  \\ \cline{2-4} 
                                        & UCM-Captions               & 9,986& optical  \\ \cline{2-4} 
                                        & RSITMD                     & 20,096& optical  \\ \cline{2-4} 
                                        & Sydney-Captions            & 2,837& optical  \\ \cline{2-4} 
                                        & NWPU-Captions              & 141,631& optical  \\ \hline
\multirow{6}{*}{\begin{tabular}[c]{@{}c@{}}VQA \\ \\ \vspace{-1.5ex} \end{tabular}} & FloodNet                    & 4,511& optical  \\ \cline{2-4} 
                                        & RSVQA$\_$LR               & 67,228& optical  \\ \cline{2-4} 
                                        & RSIVQA                     & 68,625& optical  \\ \cline{2-4} 
                                        & CRSVQA                     & 900& optical  \\ \hline
\multirow{7}{*}{Classification}         & NWPU-RESISC45              & 6,300& optical  \\ \cline{2-4} 
                                        & EuroSAT                    & 5,400& optical  \\ \cline{2-4} 
                                        & UCMerced-Landuse        & 420& optical  \\ \cline{2-4} 
                                        & WHU-RS19                   & 196& optical  \\ \cline{2-4} 
                                        & RSSCN7                     & 560& optical  \\ \cline{2-4} 
                                        & DSCR                       & 11,951& optical  \\ \cline{2-4} 
                                        & FGSCR-42                      & 3,878& optical  \\ \hline
\multirow{17}{*}{Detection}             & DOTA                  & 163,486& optical  \\ \cline{2-4} 
                                        & DIOR                       & 58,200& optical  \\ \cline{2-4} 
                                        & FAR1M                      & 40,466& optical  \\ \cline{2-4} 
                                        & NWPUVHR10                  & 3,190& optical  \\ \cline{2-4} 
                                        & HRRSD                      & 23,044& optical  \\ \cline{2-4} 
                                        & RSOD                       & 747& optical  \\ \cline{2-4} 
                                        & UCAS-AOD                       & 1,510& optical  \\ \cline{2-4} 
                                        & VisDrone                   & 186,810& optical  \\ \cline{2-4} 
                                        & AIR-SARShip-2.0              & 1,433& SAR      \\ \cline{2-4} 
                                        & SSDD                       & 1,856& SAR      \\ \cline{2-4} 
                                        & HRISD                      & 7,265& SAR      \\ \cline{2-4} 
                                        & HIT-UAV                    & 10,608& infrared \\ \cline{2-4} 
                                        & Sea-shipping            & 16,023& infrared \\ \cline{2-4} 
                                        & Infrared-security       & 17,234& infrared \\ \cline{2-4} 
                                        & Aerial-mancar           & 33,051& infrared \\ \cline{2-4} 
                                        & Double-light-vehicle & 7,922& infrared \\ \cline{2-4} 
                                        & Oceanic ship               & 2,505& infrared \\ \hline
Visual Grounding                        & DIOR-RSVG                  & 30,820& optical  \\ \hline
Region-level Captioning                        & DIOR-RSVG                  & 30,820& optical  \\ \hline
\end{tabular}
}
\end{table}

\section{Dataset Construction}
Currently, lacking domain-specific datasets for RS hampers the effective application of MLLMs in the intelligent interpretation of geographic information and open-set dialogues. Therefore, a diverse and comprehensive instruction-following dataset for image-text conversation in RS imagery is indispensable. To address this limitation, a large quantity of existing RS datasets are carefully cleaned and transformed, and a new instruction-following dataset called MMRS-1M is created covering five tasks (e.g., classification, detection, image caption, VQA, and visual grounding) and three visual modalities (e.g., optical, SAR, and infrared) is created. EarthGPT is fine-tuned on MMRS-1M to align visual and language modalities and achieve excellent coarse-grained conversation and fine-grained localization capabilities. The dataset construction process is detailed as follows.

\subsection{Coarse-grained Conversation Scenarios}
To endow EarthGPT with image-level coarse-grained question-answering capabilities, ten classification datasets, five image captioning datasets, and four VQA datasets are collected for instruction fine-tuning. 

\textbf{Category-to-instruction.} The classification datasets include the scene classification datasets AID~\cite{xia2017aid}, EuroSAT~\cite{helber2019eurosat}, 
NWPU-RESISC45~\cite{7891544}, UCMerced-LandUse~\cite{yang2010bag}, WHU-RS19~\cite{dai2010satellite}, RSSCN7~\cite{zou2015deep}, as well as the ship classification datasets FGSCR-42~\cite{di2021public} and DSCR~\cite{di2019public}. For the format conversion of the classification datasets, category-to-instruction is adopted. Specifically, the category of each image is first extracted, and the template ``What is the category of this RS image? Answering the question using a single word or phrase. Reference categories include category 1,..., category n" is used to inquire the image. The GPT model provides a response that includes the name of the category to which the image belongs.

\textbf{Caption-to-instruction.} Firstly, for image caption datasets, the existing datasets including Syndney-Captions~\cite{qu2016deep}, RSICD~\cite{lu2017exploring}, NWPU-Captions~\cite{cheng2022nwpu}, RSITMD~\cite{yuan2022exploring}, and UCM-Captions~\cite{qu2016deep} are cleaned, removing duplicate captions for the same image. Then the instruction ``Please provide a one-sentence caption for the provided RS image in detail." is provided and the EarthGPT model is tasked with describing the image. The number of captions per image determines the number of conversation rounds. Putting multiple captions for each image into a multi-turn dialogue can significantly reduce computation costs while ensuring diverse descriptions of images without compromising information leakage. This conversion process is denoted as caption-to-instruction.

\textbf{VQA-to-instruction.} For VQA datasets including Floodnet~\cite{rahnemoonfar2021floodnet}, RSVQA-LR~\cite{lobry2020rsvqa}, RSIVQA~\cite{zheng2021mutual}, and CRSVQA~\cite{zhang2023multi}, the instruction ``Answering the question using a single word or phrase" is added after the original question to control the output format of the answer. The number of question-answer pairs per image determines the number of dialogue rounds in the conversation. This multi-round dialogue allows EarthGPT to extract and interpret sufficient information from the images.

\subsection{Fine-grained Conversation Scenarios}

In order to enable the RS intelligent assistant to perform target region localization and object detection under various imaging conditions, a collection of optical, SAR, and infrared object detection datasets has been assembled. The optical datasets include DIOR~\cite{han2014object}, DOTA~\cite{9560031}, FAIR1M~\cite{sun2022fair1m}, HRRSD~\cite{zhang2019hierarchical}, NWPUVHR10~\cite{li2017rotation}, RSOD~\cite{long2017accurate}, UCAS-AOD~\cite{zhu2015orientation}, and VisDrone~\cite{zhu2021detection}. Furthermore, SAR object detection datasets like AIR-SARShip-2.0~\cite{xian2019air}, HRISD~\cite{wei2020hrsid}, SSDD~\cite{zhang2021sar} are included, along with infrared object detection datasets for HIT-UAV~\cite{suo2023hit}, Sea-shipping~\cite{Sea-shipping}, Infrared-security~\cite{Infrared-security}, Aerial-mancar~\cite{InfiRay-Aerial-mancar}, Double-light-vehicle~\cite{Double-light-vehicle} and oceanic ship~\cite{oceanic-ship}. Additionally, the DIOR-RSVG~\cite{10056343} is collected for visual grounding and region-level caption.

\textbf{Detection-to-instruction.} The detection is divided into HBB and OBB formats. The HBB format defines a bounding box as [$x_{min}$, $y_{min}$, $x_{max}$, $y_{max}$]. Here, ($x_{min}$, $y_{min}$) and ($x_{max}$, $y_{max}$) represent the corner points of the bounding box closest and farthest from the coordinate origin, respectively. On the other hand, the OBB format is defined as [$x_1$, $y_1$, $x_2$, $y_2$, $x_3$, $y_3$, $x_4$, $y_4$]. In this format, ($x_1$, $y_1$) represents the corner point of the bounding box closest to the coordinate origin, and the remaining three points are sorted in ascending order based on their angles with respect to ($x_1$, $y_1$). Additionally, for both HBB and OBB detection, the coordinates of the bounding box are normalized. During the dataset conversion process, specific instructions are used to guide the model in predicting either HBB or OBB. This process is referred to as ``detection-to-instruction". The instruction ``Detect all objects shown in the RS image and describe using horizontal bounding boxes" is used for HBB detection, while ``Detect all objects shown in the RS image and describe using oriented bounding boxes" is used for OBB detection. The model then outputs the coordinates and categories of all objects in the image. Furthermore, to achieve referring detection, an additional multi-round conversation session is incorporated to detect objects of each category.

\textbf{VG-to-instruction.} For the visual grounding dataset including DIOR-RSVG, two types of conversation formats are constructed via VG-to-instruction. Specifically, in the first format, users describe the region of interest and provide instructions for MLLM to locate the corresponding target. The model then outputs the coordinates of the identified target. The second format involves the user inputting instructions to describe the target region based on its spatial location coordinates, and MLLM generates a description of the corresponding target region. The former format bestows EarthGPT with the ability to perform visual localization, while the latter is employed to accomplish the region caption task. 

The aforementioned is a detailed description of the MMRS-1M dataset construction process. Additionally, for the datasets mentioned above, the training set and validation are processed through data transformation and the construction of MMRS-1M after image cutting and dataset splitting. The details on the training samples can be found in Tab. \ref{tab:MMRS-1M Compoents}. The construction flow, as depicted in Fig. \ref{FIG:datasets}, provides a more concrete and intuitive understanding of the dataset construction process. 

In summary, we clean and convert 34 various existing RS datasets into a uniform visual instruction-following format, covering optical, SAR, and infrared visual modalities, facilitating the application of MLLMs in the RS domain.

\section{Experiments and Analysis}
In this section, we conduct extensive experiments to validate the performance of our proposed EarthGPT. In subsection A, we describe the implementation details. Subsequently, we demonstrate EarthGPT's powerful ability in various scenarios such as classification, image captioning, VQA, visual grounding, and object detection, compared to other MLMs, specialist models, and open-set models.

\subsection{Implementation Details}
In the visual-enhanced perception stage, the DINOv2 ViT-L/14~\cite{oquab2023dinov2} is adopted as the ViT encoder, and the frozen CLIP ConvNeXt-L~\cite{ilharco_gabriel_2021_5143773} is utilized as the CNN encoder. The two visual encoders are kept frozen throughout the training. Subsequently, in the cross-modal mutual comprehension stage, the objective is to convert language-only LLM into an MLLM.  We trained EarthGPT on LAION-400M\cite{schuhmann2021laion}, COCO Caption\cite{chen2015microsoft} datasets, which primarily focus on natural scene image captioning, to develop basic multi-modal alignment from scratch. Furthermore, in the unified multi-task tuning phase, the goal is to equip MLLM with the versatility needed for diverse RS downstream tasks. As mentioned in Section IV, the MMRS-1M dataset is constructed based on diverse and comprehensive RS task-specific datasets. By unifying all the datasets into a multi-modal conversational format, the training process is optimized, reducing costs and enhancing efficiency. During the training, we only train an off-the-shelf language model LLaMA-2\cite{touvron2023llama_b} and randomly initialized visual projections. We use the AdamW optimizer $ (\beta 1, \beta 2) = (0.9, 0.95)$, a maximum learning rate of $2\times 10^{-5}$, a minimum learning rate of 0. 

\subsection{Scene Classification}

\begin{table}[!b]
\centering % 居中设置
\caption{Supervised comparison results on NWPU-RESISC45 for specialist models and our EarthGPT. }\label{tab:Classification_compare_supervised}
\renewcommand{\arraystretch}{1.2}
\scalebox{1}{
\fontsize{9pt}{9pt}\selectfont
%\small
%\setlength{\tabcolsep}{2pt}
\begin{tabular}{lc|c}
\toprule
\multicolumn{1}{c}{Method}                 & \multicolumn{1}{l|}{Publication Year} & Top-1 Acc  \\ 
\cmidrule(lr){1-2}\cmidrule(lr){3-3}
\textit{\textbf{Specialist Models}} &                  &            \\ 
\multicolumn{1}{l}{SeCo~\cite{manas2021seasonal}}                      & \multicolumn{1}{l|}{ICCV 2021}         & 92.71      \\
\multicolumn{1}{l}{MGSNet~\cite{10184498}}                       & \multicolumn{1}{l|}{TGRS 2023}         & \textbf{94.57}      \\
\multicolumn{1}{l}{CSDS~\cite{wang2021csds}}                       & \multicolumn{1}{l|}{JSTARS 2021}        & 93.59      \\
\multicolumn{1}{l}{T-CNN~\cite{wang2022transferring}}                      & \multicolumn{1}{l|}{TGRS 2022}         & 93.05      \\
\multicolumn{1}{l}{PSGAN~\cite{cheng2021perturbation}}                      & \multicolumn{1}{l|}{TGRS 2022}         & 88.47      \\ 
\cmidrule(lr){1-2}\cmidrule(lr){3-3}
\textit{\textbf{MLLM}}              &                  &            \\ 
\rowcolor[RGB]{248,240,205}\multicolumn{1}{l} {\textbf{EarthGPT(Ours)}}      & \multicolumn{1}{l|}{}      & {93.84} \\ 
\bottomrule
\end{tabular}
} 
\end{table}

\begin{table}[!b]
\caption{Zero-shot comparison results on CLRS and NaSC-TG2 for other MLLMs and our EarthGPT. }\label{tab:Classification_compare_zeroshot}
\renewcommand{\arraystretch}{1.5}
\scalebox{0.9}{
\fontsize{10pt}{10pt}\selectfont
\begin{tabular}{lccc}  
\toprule      
\multicolumn{1}{c}{Method}          & \multicolumn{1}{l|}{Publication Year} & CLRS & NaSC-TG2  \\ 
\cmidrule(lr){1-2}\cmidrule(lr){3-4}
\multicolumn{1}{l}{Qwen-VL-Chat~\cite{bai2023qwen}} & \multicolumn{1}{l|}{Arxiv 2023}       & 51.76& 36.09                 \\
\multicolumn{1}{l}{LLaVa-1.5~\cite{liu2023visual}}    & \multicolumn{1}{l|}{NeurIPS 2023}     & 55.86    & 40.67                  \\
\multicolumn{1}{l}{Sphinx~\cite{lin2023sphinx}}      & \multicolumn{1}{l|}{Arxiv 2023}       & 60.72   & 49.76                  \\
\cmidrule(lr){1-2}\cmidrule(lr){3-4}
\rowcolor[RGB]{248,240,205} \multicolumn{1}{l}{\textbf{EarthGPT(Ours)}} &\multicolumn{1}{l|}{}                 & \textbf{77.37}    & \textbf{74.72}          \\ 
\bottomrule
\end{tabular}
} 
\end{table}

To evaluate the classification performance, supervised and zero-shot classification assessments are conducted. In the first evaluation, the test sets of NWPU-RESISC45 are utilized. The NWPU-RESISC45 dataset is a large-scale publicly available dataset for RS image scene classification, published by Northwestern Polytechnical University. It is comprised of 45 scene categories, with each category containing 700 RS images. Each image has a size of 256$\times$256 pixels. 80$\%$ of the dataset is used as the test set to evaluate and compare EarthGPT with other specialist models.

For zero-shot classification, all the images from the CLRS~\cite{li2020clrs} and NaSC-TG2~\cite{zhou2021nasc} datasets are employed as the zero-shot testing set. The CLRS dataset includes 15,000 RS images with 25 land-use types such as airports, beaches, etc. These images were extracted from Google Earth, Bing Maps, Google Maps, and Tianditu, and were released by Central South University in 2020. Each image has a size of 256$\times$256$\times$3, and the resolutions range from 0.26 m to 8.85 m. The NaSC-TG2 dataset consists of 20,000 RS images of 10 land cover types. These images were extracted from the Tiangong-2 satellite and were released by the Center for Space Applications and Engineering of the Chinese Academy of Sciences in 2021. Each image in this dataset has a size of 128$\times$128$\times$3, and the pixel resolution is 100 m. For the evaluation of supervised classification and zero-shot classification, Top-1 accuracy is reported. 

In the classification evaluation, similar to the training phase, we present all categories from the datasets as reference categories. The model is instructed to classify images using only one word or phrase. First, we report the accuracy for EarthGPT on the test sets of NWPU-RESISC45 in Tab. \ref{tab:Classification_compare_supervised}, comparing its performance with some SOTA specialist models. It is observed that EarthGPT outperforms SOTA specialist models such as CSDS, and T-CNN with improvements of 0.25$\%$ and 0.79$\%$, achieving comparable performance with MGSNet. In the zero-shot classification evaluation, we compare EarthGPT with other MLLMs such as QwenVL-Chat, LLaVa-1.5, and Sphinx in Tab. \ref{tab:Classification_compare_zeroshot}. It is apparent that EarthGPT shows a notable improvement in the zero-shot evaluation compared to other MLLMs. Specifically, EarthGPT outperforms Qwen-VL-Chat, LLavaV1.5, and Sphinx in top-1 accuracy on the CLRS dataset, achieving 25.61$\%$, 21.51$\%$, 16.65$\%$ respectively. Additionally, EarthGPT brings improvements of 38.63$\%$, 34.05$\%$, and 24.96$\%$ on NaSC-TG2 dataset compared with Qwen-VL-Chat, LLavaV1.5, and Sphinx, respectively. This suggests that the integration of domain knowledge from RS is highly important for generalizing to unknown classification scenarios.

\begin{table*}[!h]
\caption{Supervised comparison results on NWPU caption for specialist models and our MLLM EarthGPT. }\label{tab:NWPU_caption _compare_supervised}
\renewcommand{\arraystretch}{1.5}
\scalebox{1.1}{
\begin{tabular}{lccccccccc}
\toprule
\multicolumn{1}{c}{Method} & \multicolumn{1}{c|}{Publication Year}& BLEU1 & BLEU2 & BLEU3 & BLEU4 & METEOR & ROUGE\_L & CIDEr(0-5) & SPICE \\ 
\cmidrule(lr){1-2} \cmidrule(lr){3-10}
\textit{\textbf{Specialist Model}}        &    &       &       &       &       &        &         &            &       \\ 
\multicolumn{1}{l}{CSMLF~\cite{8633358}}                 & \multicolumn{1}{l|}{GRSL 2019}     & 77.0 & 64.9 & 53.2 & 47.1 & 32.0  & 57.8   & 106.5      & 26.5 \\
\multicolumn{1}{l}{Multimodal~\cite{qu2016deep}}           & \multicolumn{1}{l|}{CITS 2016}    & 72.5 & 60.3 & 51.8 & 45.5 & 33.6  & 59.1   & 117.9      & 27.6 \\
\multicolumn{1}{l}{Attention(soft)~\cite{lu2017exploring}}      & \multicolumn{1}{l|}{TGRS 2017}  & 73.1 & 60.9 & 52.5 & 46.2 & 33.9  & 59.9   & 113.6      & 28.5 \\
\multicolumn{1}{l}{Attention(hard)~\cite{lu2017exploring}}       & \multicolumn{1}{l|}{TGRS 2017}    & 73.3 & 61.0 & 52.7 & 46.4 & 34.0  & 60.0   & 110.3      & 28.4 \\
\multicolumn{1}{l}{FC-Att~\cite{zhang2019description}}           & \multicolumn{1}{l|}{TGRS 2019}   & 73.6 & 61.5 & 53.2 & 46.9 & 33.8  & 60.0   & 123.1      & 28.3 \\
\multicolumn{1}{l}{SM-Att~\cite{zhang2019description}}           & \multicolumn{1}{l|}{TGRS 2019}    & 73.9 & 61.7 & 53.2 & 46.8 & 33.0  & 59.3   & 123.6      & 27.6 \\
\multicolumn{1}{l}{MLCA-Net~\cite{9866055}}              & \multicolumn{1}{l|}{TGRS 2022}    & 74.5 & 62.4 & 54.1 & 47.8 & 33.7  & 60.1   & 126.4      & 28.5 \\ 
\cmidrule(lr){1-2} \cmidrule(lr){3-10}
\textit{\textbf{MLLM}}                 \\ 
 \rowcolor[RGB]{248,240,205} \multicolumn{1}{l} {\textbf{EarthGPT(Ours)}} & \multicolumn{1}{l|}{} & \textbf{87.1} & \textbf{78.7} & \textbf{71.6} & \textbf{65.5} & \textbf{44.5} & \textbf{78.2} & \textbf{192.6} & \textbf{32.2} \\

\bottomrule
\end{tabular}
}
\end{table*}

\subsection{Image Captioning}
For the evaluation of image captioning capability, the test set of the NWPU-Caption dataset is utilized to assess and compare EarthGPT with specialist models in supervised setting. The NWPU-Caption dataset is created by Northwestern Polytechnical University, incorporating 31,500 aerial RS images and 157,500 sentences for RS image caption. Following the setting of MLCA-Net, we employ BLEU\-1, BLEU\-2, BLEU\-3, BLEU\-4, METEOR, ROUGE\-L, and CIDEr\-D as evaluation metrics. 

From Tab. \ref{tab:NWPU_caption _compare_supervised}, it can be clearly seen that on the NWPU-Captions dataset, compared to other SOTA methods, EarthGPT improves 10.1$\%$, 13.8$\%$, 17.5$\%$, 17.7$\%$, 10.5$\%$, 18.2$\%$, 66.2$\%$, and 3.7$\%$ in terms of BLEU-1, BLEU-2, BLEU-3, BLEU-4, METEOR, ROUGE-L, and CIDEr-D. It is evident that EarthGPT provides accurate, detailed, and diverse descriptions of RS images.

\begin{table}[]
\centering 
\caption{Supervised comparison results on CRSVQA for specialist models and our EarthGPT. }\label{tab:CRSVQA_compare_supervised}
\renewcommand{\arraystretch}{1.3}
\scalebox{1.2}{
\fontsize{9pt}{9pt}\selectfont
%\small
%\setlength{\tabcolsep}{2pt}
\begin{tabular}{lcc}
\toprule
\multicolumn{1}{c}{Method}                  & \multicolumn{1}{l|}{Publication Year} & OA  \\ 
\cmidrule(lr){1-2}\cmidrule(lr){3-3}
\textit{\textbf{Specialist models}} &                  &            \\ 
\multicolumn{1}{l} {Qonly~\cite{marino2019ok}}                   & \multicolumn{1}{l|}{CVPR 2019}         & 23.49     \\
\multicolumn{1}{l}{RSVQA~\cite{lobry2020rsvqa}}                      & \multicolumn{1}{l|}{TGRS 2020}           & 58.96      \\
\multicolumn{1}{l}{RSVQA(GRU)~\cite{lobry2020rsvqa}}                       & \multicolumn{1}{l|}{TGRS 2020}           & 59.41      \\
\multicolumn{1}{l}{SAN~\cite{kafle2016answer}}                       & \multicolumn{1}{l|}{CVPR 2016}          & 61.17      \\
\multicolumn{1}{l}{MQVQA~\cite{zhang2023multi}}                      & \multicolumn{1}{l|}{TGRS 2023}           & 70.91      \\
\cmidrule(lr){1-2}\cmidrule(lr){3-3}
\textbf{\textit{MLLM}}            {}                   &            \\ 
\rowcolor[RGB]{248,240,205} \multicolumn{1}{l} {{\textbf{EarthGPT(Ours)}} }     & \multicolumn{1}{l|}{}     & {\textbf{82.00} }\\ 

\bottomrule
\end{tabular}
} % adjustbox
\end{table}
\begin{table}[]
\caption{Zero-shot comparison results on RSVQA-HR for other MLLMs and our  EarthGPT.}\label{tab:RSVQA_high_compare_zeroshot}
\renewcommand{\arraystretch}{1.5}
\scalebox{0.9}{
\fontsize{9.5pt}{9.5pt}\selectfont
\begin{tabular}{lcccc}  
\toprule      
\multicolumn{1}{c}{Method}\!\!\!\!          & \multicolumn{1}{c|}{\!\!\!\!\!\!\!\!Publication Year} & \!\!Presence\!\! \!\!& \!\!Comparison\!\! \!\!&\!\! \!\!OA \\ 
\cmidrule(lr){1-2}\cmidrule(lr){3-5}
\multicolumn{1}{l}{Qwen-VL-Chat~\cite{bai2023qwen}} &  \multicolumn{1}{l|}{Arxiv 2023}       & \textbf{69.83 }& 67.29        & 68.40          \\
\multicolumn{1}{l}{LLaVa-1.5~\cite{liu2023visual}}    & \multicolumn{1}{l|}{NeurIPS 2023}     & 66.44    & 60.41        & 63.06          \\
\multicolumn{1}{l}{MiniGPTV2~\cite{chen2023minigpt}}      & \multicolumn{1}{l|}{Arxiv 2023}       & 40.79   &50.91        & 46.46          \\
\multicolumn{1}{l}{Sphinx~\cite{lin2023sphinx}}      & \multicolumn{1}{l|}{Arxiv 2023}       & 64.28   &74.75        & 69.79          \\
\multicolumn{1}{l}{GeoChat~\cite{kuckreja2023geochat}}      & \multicolumn{1}{l|}{Arxiv 2023}       & 58.45   &\textbf{83.19 }       & \textbf{72.30}          \\
\cmidrule(lr){1-2}\cmidrule(lr){3-5}
\rowcolor[RGB]{248,240,205} \multicolumn{1}{l}
{{\textbf{EarthGPT(Ours)}} }                  & \multicolumn{1}{l|}{}    & {62.77}  & {79.53}      & {72.06}         \\ 
\bottomrule
\end{tabular}
} % adjustbox

\end{table}

\begin{table*}[!t]
\caption{Supervised comparison results on DIOR-RSVG for specialist models, other MLLMs, and our  EarthGPT.}\label{tab:DIOR-RSVG_compare_supervised}
\renewcommand{\arraystretch}{1.2}
\scalebox{1.3}{
\begin{tabular}{lccccccccc}
\toprule
\multicolumn{1}{c}{Method} & \multicolumn{1}{c|}{Publication Year}& Pr@0.5 & Pr@0.6 & Pr@0.7 & Pr@0.8 & Pr@0.9 & mIoU & cIoU  \\ 
\cmidrule(lr){1-2} \cmidrule(lr){3-9}
\textit{\textbf{Specialist Model}}        &    &       &       &    &       &       &           \\ 
\multicolumn{1}{l}{ZSGNet~\cite{sadhu2019zero}}                 & \multicolumn{1}{l|}{ICCV 2019}     & 51.67 & 48.13 & 42.30 & 32.41  & 10.15 & 44.12 & 51.65\\
\multicolumn{1}{l}{FAOA~\cite{yang2019fast}}           & \multicolumn{1}{l|}{ICCV 2019}   & 70.86 & 67.37 & 62.04 & 53.19  & 36.44 & 62.86 & 67.28\\
\multicolumn{1}{l}{ReSC~\cite{yang2020improving}}      & \multicolumn{1}{l|}{ECCV 2020}  & 72.71 & 68.92 & 63.01 & 53.70 & 33.37 & 64.24 & 68.10  \\
\multicolumn{1}{l}{LBYL-Net~\cite{huang2021look}}       & \multicolumn{1}{l|}{CVPR 2021}    & 73.78 & 69.22 & 65.56 & 47.89 & 15.69  & 65.92 & 76.37 \\
\multicolumn{1}{l}{TransVG~\cite{deng2021transvg}}       & \multicolumn{1}{l|}{ICCV 2021}     & 72.41 & 67.38 & 60.05 & 49.10 & 27.84  & 63.56 & 76.27 \\
\multicolumn{1}{l}{VLTVG\cite{yang2022improving}}       & \multicolumn{1}{l|}{CVPR 2021}     & 75.79 & 72.22 & 66.33 & 55.17 & 33.11  & 66.32 & 77.85 \\
\multicolumn{1}{l}{MGVLF~\cite{10056343}}       & \multicolumn{1}{l|}{TGRS 2023}     & \textbf{76.78} & \textbf{72.68} & \textbf{66.74} & 56.42 & 35.07  & 68.04 & 78.41 \\
\cmidrule(lr){1-2} \cmidrule(lr){3-9}
\textit{\textbf{MLLM}}                   &     &       &       &    &       &       &        \\ 
    \rowcolor[RGB]{248,240,205}  \multicolumn{1}{l}{\textbf{EarthGPT(Ours)}} & \multicolumn{1}{l|}{} & 76.65 & 71.93 & 66.52 & \textbf{56.53} & \textbf{37.63} & \textbf{69.34} & \textbf{81.54} \\

\bottomrule
\end{tabular}
}
\end{table*}

\subsection{Visual Question Answering}
The evaluation of VQA is also divided into supervised and zero-shot assessments. For the supervised VQA setting, we adopt the test set of CRSVQA including 1000 image-question-answer pairs. Following the setting of MQVQA, 10$\%$ of image-question-answer pairs are adopted as the test set, and the overall accuracy (OA) is reported. For zero-shot VQA setting, RSVQA-HR is utilized to evaluate the performance of EarthGPT and other MLLMs. The RSVQA-HR dataset was created from high-resolution orthophoto (HRO) datasets obtained from USGS. It consists of 10,659 images, with 1,066,316 question-answer pairs. RSVQA-HR dataset is divided into training, validation, test set 1, and test set 2. In the evaluation of zero-shot VQA, test set 2 is adopted for evaluating the model's robustness to diverse locations. Additionally, following the setting of RSGPT and GeoChat, object counting problems are not included in the computation of metrics for RSVQA-HR. For the RSVQA-HR dataset, the accuracy of different question categories and overall accuracy have been reported.

From Tab. \ref{tab:CRSVQA_compare_supervised}, it is observed that EarthGPT significantly outperforms other specialist models in terms of OA on CRSVQA datasets. Specifically, EarthGPT surpasses MQVQA and SAN with improvements of 11.09$\%$ and 20.83$\%$, respectively. In zero-shot evaluation, EarthGPT achieves an average accuracy of 72.05$\%$ on RSVQA-HR above or commensurate with other MLLMs, as shown in Tab. \ref{tab:RSVQA_high_compare_zeroshot}. Compared with Qwen-VL-Chat, LLava v1.5, minigptV2, and Sphinx, EarthGPT brings the improvement of 3.66$\%$, 9.00$\%$, 25.60$\%$, and 2.27$\%$, respectively. In conclusion, EarthGPT demonstrates exceptional capabilities in answering questions within entirely new and unfamiliar RS contexts, and a profound ability to comprehend the true significance of queries presented in visual data and text instructions.

\subsection{Visual Grounding}
To evaluate the visual grounding performance, we use the test set of DIOR-RSVG containing 758 grounding questions. DIOR-RSVG dataset comprises 38,320 language expressions across 17,402 RS images, uniquely characterizing individual objects within 20 diverse categories. Evaluation metrics include Pr@0.5, Pr@0.6, Pr@0.7, Pr@0.8, Pr@0.9 mean IoU(mIoU), and cum IoU(cIoU). Specialist models ZSGNet, FAOA, ReSc, LBYL-Net, TranVG, VLTVG and MGVLF are adopted to compare with EarthGPT. 

Tab. \ref{tab:DIOR-RSVG_compare_supervised} shows the performance of EarthGPT and other specialist models on the DIOR-RSVG test set. It can be seen that EarthGPT brings significant performance improvements of 0.11$\%$, 1.19$\%$, 1.30$\%$, and 3.13$\%$ on Pr@0.8, Pr@0.9, mIoU and cIoU, respectively. It is evident that EarthGPT have exceptional spatial location and region-level image comprehension ability.

\begin{table}[!h]
\caption{Zero-shot comparison results on MAR20 for other methods and our  EarthGPT.}\label{tab:MAR20_compare_zeroshot}
\renewcommand{\arraystretch}{1.5}
\scalebox{0.9}{
\fontsize{10.2pt}{10.2pt}\selectfont
\begin{tabular}{lccc} % 
\toprule
\multicolumn{1}{c}{\!\!\!\! Method\!\!\!\!\!\!\!\!\!\!\!\!}& \multicolumn{1}{c|}{ \!\!\!\!\!\!Publication Year} & \!\! AP@40 &\!\!\!\!\! AP@50 \\ 
\cmidrule(lr){1-2}\cmidrule(lr){3-4}
\textit{\textbf{Open-set Model}}          \\ 
\multicolumn{1}{l}{GroundingDINO~\cite{liu2023grounding}} &  \multicolumn{1}{l|}{Arxiv 2023}       & 88.55 & 88.21              \\
\multicolumn{1}{l}{mm-GroundingDINO~\cite{zhao2024open}}    & \multicolumn{1}{l|}{Arxiv 2024}     & 88.37    & 88.14           \\
\cmidrule(lr){1-2}\cmidrule(lr){3-4}
\textit{\textbf{MLLM}}          \\ 
\multicolumn{1}{l}{Lenna~\cite{wei2023lenna}}      & \multicolumn{1}{l|}{Arxiv 2023}       & 72.69   &72.11      \\
\multicolumn{1}{l}{Qwen-VL-Chat~\cite{bai2023qwen}}      & \multicolumn{1}{l|}{Arxiv 2023}       & 72.56   &40.04     \\
\multicolumn{1}{l}{Sphinx~\cite{lin2023sphinx}}      & \multicolumn{1}{l|}{Arxiv 2023}       & 81.03  &80.48          \\
\rowcolor[RGB]{248,240,205}\multicolumn{1}{l}{{\textbf{EarthGPT(Ours)}}}                  & \multicolumn{1}{l|}{}    & {\textbf{90.47}}  & {\textbf{90.11 }}          \\ 
\bottomrule
\end{tabular}
} % adjustbox
\end{table}
%\vspace{-1cm}
\begin{table}[!h]
\caption{Zero-shot comparison results for OBB detection on MAR20 for other methods and our  EarthGPT.}\label{tab:MAR_compare_zeroshot_obb}
\renewcommand{\arraystretch}{1.5}
\scalebox{0.9}{
\fontsize{10.2pt}{10.2pt}\selectfont
\begin{tabular}{lccc} % 删除了多余的列，只保留需要的列
\toprule
\multicolumn{1}{c}{\!\!\!\! Method\!\!\!\!\!\!\!\!\!\!\!\!}& \multicolumn{1}{c|}{ \!\!\!\!\!\!Publication Year} & \!\! AP@40 &\!\!\!\!\! AP@50 \\ 
\cmidrule(lr){1-2}\cmidrule(lr){3-4}
\textit{\textbf{Specialist Model}}     \\ 
\multicolumn{1}{l}{S2A-Net~\cite{han2021align}}      & \multicolumn{1}{l|}{TGRS 2021}       & 90.71  &90.23\\
\multicolumn{1}{l}{CFA~\cite{guo2021beyond}}      & \multicolumn{1}{l|}{CVPR 2021}       & 90.66  &90.28 \\
\multicolumn{1}{l}{Oriented RepPoints~\cite{li2022oriented}}      & \multicolumn{1}{l|}{CVPR 2022}       & 90.68  &\textbf{90.56} \\
\multicolumn{1}{l}{Oriented R-CNN~\cite{xie2021oriented}}      & \multicolumn{1}{l|}{ICCV 2021}       & \textbf{90.70}  &90.54 \\
\multicolumn{1}{l}{Sasm~\cite{hou2022shape}}      & \multicolumn{1}{l|}{AAAI 2022}       & 90.19 &89.83 \\
\multicolumn{1}{l}{R3Det~\cite{yang2021r3det}}      & \multicolumn{1}{l|}{AAAI 2021}       & 90.61  &90.11 \\
\cmidrule(lr){1-2}\cmidrule(lr){3-4}
\textit{\textbf{MLLM}}            \\ 
\rowcolor[RGB]{248,240,205}\multicolumn{1}{l}{\textbf{EarthGPT(Ours)}}                 & \multicolumn{1}{l|}{}    & {{90.53}}  & {{87.86 }} \\
\bottomrule
\end{tabular}
} % adjustbox
\end{table}

\subsection{Object Detection}

To thoroughly assess the potential and generalization capabilities of the innovative language-guided paradigm of EarthGPT in object detection, we adopt zero-shot setting for comparison with other MLLMs and specialist models. Given that other MLLMs are primarily designed for HBB detection and struggle with OBB detection, we utilize the HBB format of the MAR20 dataset~\cite{yu2022mar20} to evaluate the detection performance, employing AP@40 and AP@50 as metrics. The MAR20 dataset is a large-scale RS aircraft target recognition dataset. It comprises 3,842 images with 22,341 instances, and each target instance is annotated with both horizontal bounding boxes and oriented bounding boxes. Comparing methods include MLLMs Qwen-VL-Chat, Sphinx, Lenna, and open-set object detection models GroundingDINO and mm-GroundingDINO. To address the challenge of MLLMs not predicting confidence scores, we employ clip-score as a confidence logit for noise filtering. Remoteclip~\cite{liu2023remoteclip} weights are adopted to compute the clipscore. For the evaluation of OBB detection, we adopt the OBB format of MAR20 to compare with specialist models trained on DOTA, such as S2A$\_$Net, CFA, Oriented RepPoints, Oriented R-CNN, Sasm, and R3Det.

From Tab. \ref{tab:MAR20_compare_zeroshot}, on the test set of MAR20, EarthGPT exhibits notable enhancements of AP@40, AP@50 compared to other MLLMs and open-set detection models, underscoring its robust generalization capability under unseen scenarios. Specifically, EarthGPT brings improvements of 1.92$\%$ and 1.90$\%$ in terms of AP@40 and AP@50 compared with the open-set model GroundingDINO. For OBB detection, Tab. \ref{tab:MAR_compare_zeroshot_obb} demonstrates that EarthGPT achieves competitive performance with specialist models trained on DOTA, showcasing the powerful potential of language-based generative detection paradigm.

\section{Visualization}

In this section, we present the qualitative experimental result of EarthGPT to demonstrate proficiency in multi-turn dialogue and diverse RS tasks under multi-sensor visual modalities. EarthGPT showcases a remarkable capacity for image-level and region-level visual perception, and adeptly interpreting RS complex visual data. Notably, EarthGPT also shows advanced performance in chain-of-thought reasoning, fostering the emergence of cross-task visual comprehension. 

\subsection{Multi-turn Multi-task Dialogue}
Fig. \ref{FIG:Multi-turn dialogue} provides a detailed depiction of EarthGPT's capabilities in conducting multi-turn dialogues. Diverging from traditional specialist models limited to specific tasks, EarthGPT leverages language as a medium to achieve diverse abilities in multi-turn dialogue, including classification, image captioning, VQA, object detection, region-level captioning, and visual grounding. In Fig. \ref{FIG:Multi-turn dialogue}, we demonstrate EarthGPT's proficiency in accurately identifying the scene as a ground track field, along with a detailed spatial layout description. EarthGPT localizes and describes specific areas precisely, such as the ground track field and the nearby small baseball field. Compared to the previous specialist models, EarthGPT enhances readability and comprehension, making it more user-friendly and accessible.

\begin{figure*}[!t]
	\centering
		\includegraphics[scale=0.13]{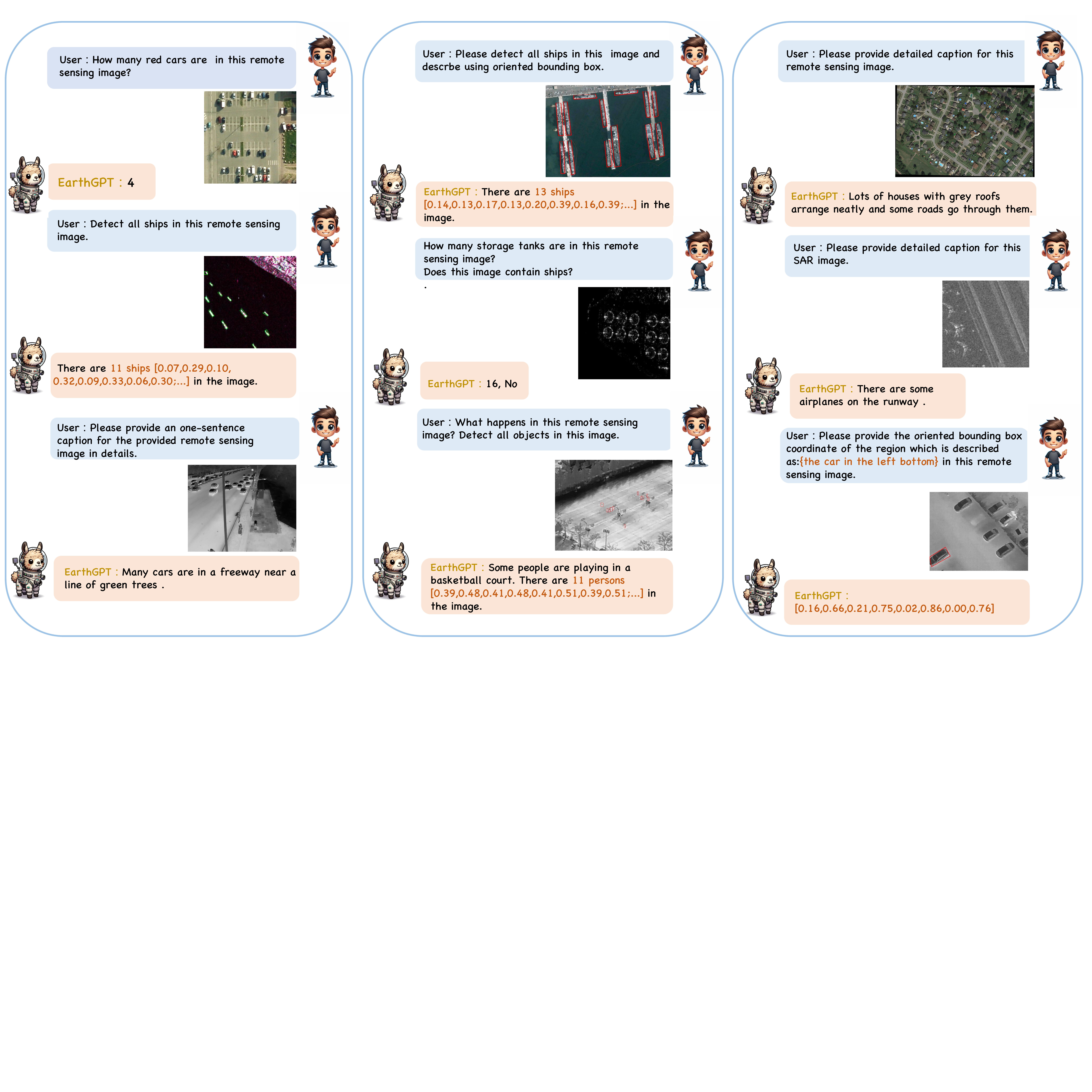}
	\caption{Examples of EarthGPT for different visual modalities inference ability.}
	\label{FIG:Multi-vision }
\end{figure*}

\begin{figure*}[!t]
	\centering
		\includegraphics[scale=0.122]{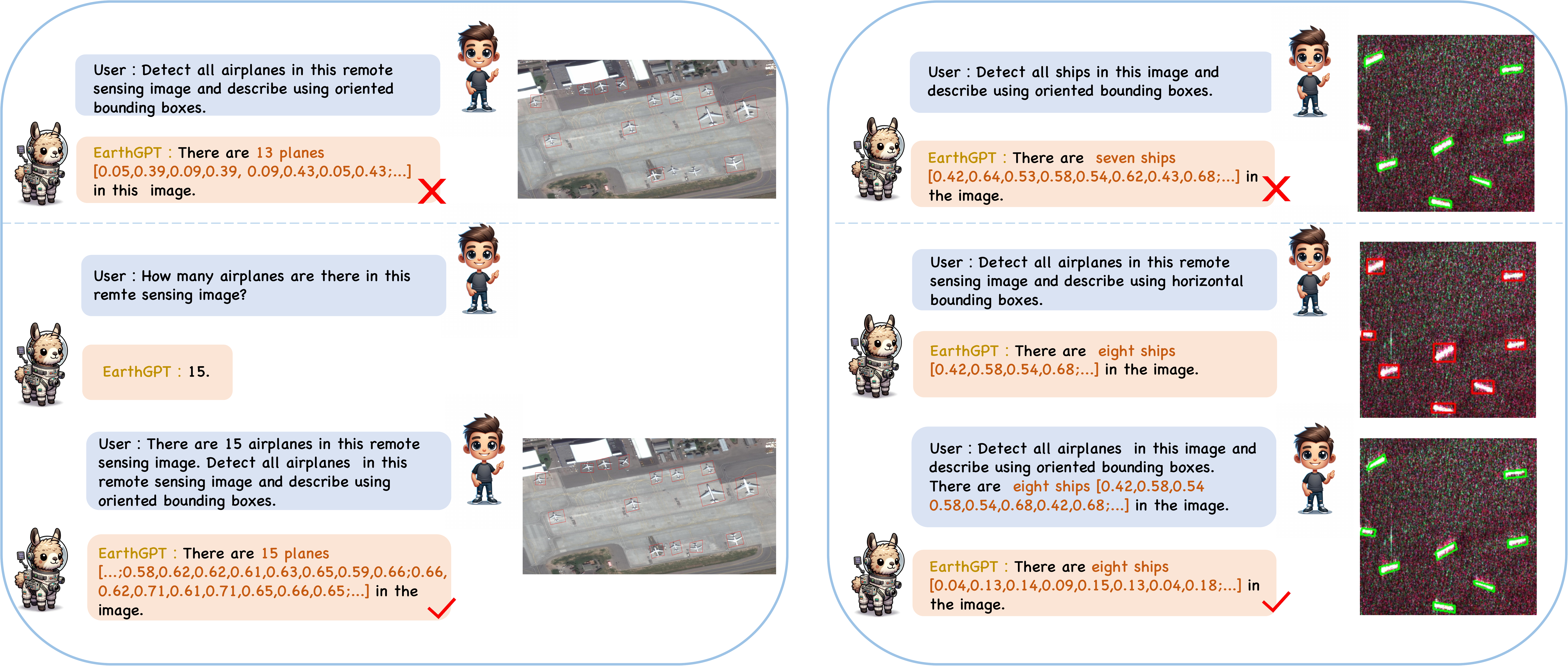}
	\caption{Examples of the chain-of-thought prompting for EarthGPT to perform visual reasoning.}
	\label{FIG:Cross_task}
\end{figure*}
\begin{figure*}[!t]
	\centering
		\includegraphics[scale=0.140]{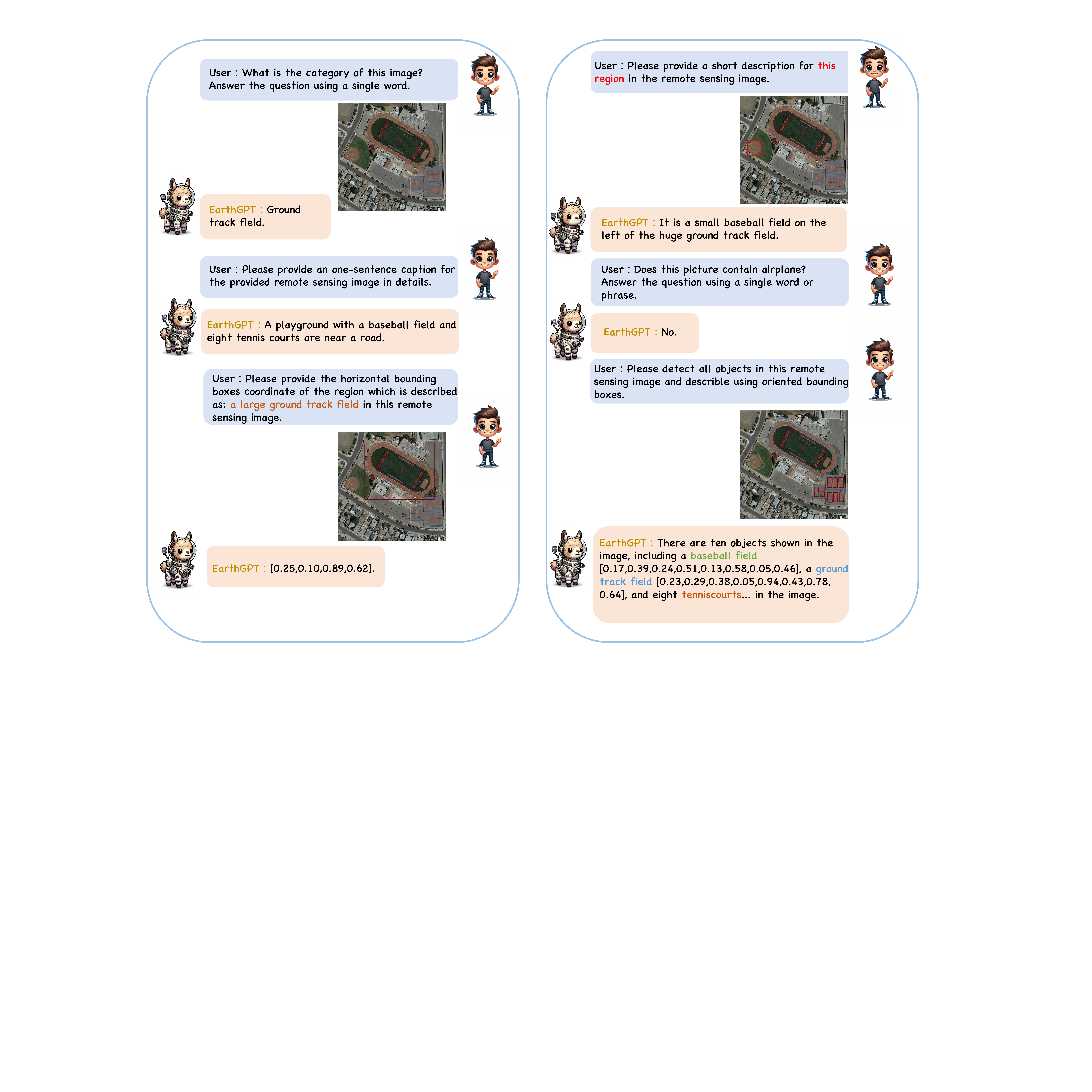}
	\caption{Examples of EarthGPT for multi-turn various RS tasks dialogue.}
	\label{FIG:Multi-turn dialogue}
\end{figure*}
\subsection{Multi Visual Modality Inference}
Previous specialist models in the RS domain mainly concentrate on single visual modality and single task. Thanks to our MMRS-1M's rich multi-sensor information, laying the foundation for training the versatile EarthGPT. EarthGPT has robust generalization capabilities across optical, SAR, and infrared visual modalities. Fig. \ref{FIG:Multi-vision } showcases EarthGPT's remarkable performance across a variety of tasks in different visual modalities.

Although the MMRS-1M dataset involves limited SAR and infrared data compared to the optical data. Note that EarthGPT overcomes the constraints by using language to integrate different visual modalities. As shown in Fig. \ref{FIG:Multi-vision }, in optical imagery, EarthGPT can accurately enumerate the number of red cars in the scene, identify all ships on water surfaces, and provide detailed descriptions of dense residential.
In SAR imagery, EarthGPT accurately counts storage tanks, evaluates ship presence, and concisely describes the airport. In infrared imagery, EarthGPT precisely locates cars in certain areas, identifies persons on a basketball court, and accurately interprets their activities. Those instances illustrate EarthGPT's exceptional knowledge transfer and generalization abilities.

\subsection{Chain-of-thought Prompting for Visual Reasoning}

We discover that EarthGPT can employ chain-of-thought prompting to enhance visual understanding and reasoning accuracy. For instance, EarthGPT can enhance the OBB detection completeness through the result of object counting as the prompt.  As shown in Fig. \ref{FIG:Cross_task}, in the airport detection scene, EarthGPT initially misses two planes, when promptly by the question ``How many planes are there in the airport?", after discovering there are 15 planes in total, EarthGPT intelligently uses this newfound knowledge to re-evaluate the aircraft detection task, ultimately successfully identifying all the planes. In addition, EarthGPT is adept at enhancing the accuracy of OBB detection by harnessing HBB detection results. In the scenario involving SAR ships, EarthGPT effectively detects all ships using HBB but misses ships using OBB. To improve the OBB results, EarthGPT utilizes the HBB detection results as a valuable hint to rectify the omissions and deliver more precise OBB results. In conclusion, the chain-of-thought prompting technique facilitates intelligently incorporating insights from related tasks to elevate the performance in various visual interpretation and reasoning tasks.

\section{Conclusion}
In this paper, a versatile MLLM EarthGPT, which unifies a wide range of RS tasks and various multi-sensor images, has been proposed for universal RS image comprehension. EarthGPT integrates coarse-scale and fine-scale visual perception information, bridging the gaps in cross-modal mutual comprehension and vision reasoning. More importantly, EarthGPT is capable of multi-sensor image interpretation and RS downstream tasks including scene classification, image captioning, region-level captioning, VQA, visual grounding, object detection. Furthermore, the MMRS-1M dataset, a multi-sensor multi-modal RS instruction-following dataset comprising more than 1M image-text pairs, has been constructed. The MMRS-1M dataset tackles the limitation of MLLMs on RS expert knowledge and encourages the growth of MLLMs tailored specifically for applications in the RS domain. Extensive experiments have demonstrated that EarthGPT surpasses the numerous existing specialist models and MLLMs in various RS visual interpretation tasks, and provides an open-set reasoning capability suitable for multiple RS downstream tasks, both in supervised and zero-shot settings. In the future, we will focus on improving the OBB detection performance of MLLMs and incorporating more modalities into EarthGPT for all-purpose capabilities.

\bibliographystyle{unsrt}
\bibliography{my.bib}

\begin{thebibliography}{100}

\bibitem{ma2019deep}
Lei Ma, Yu~Liu, Xueliang Zhang, Yuanxin Ye, Gaofei Yin, and Brian~Alan Johnson.
\newblock Deep learning in remote sensing applications: A meta-analysis and review.
\newblock {\em ISPRS journal of photogrammetry and remote sensing}, 152:166--177, 2019.

\bibitem{zhu2017deep}
Xiao~Xiang Zhu, Devis Tuia, Lichao Mou, Gui-Song Xia, Liangpei Zhang, Feng Xu, and Friedrich Fraundorfer.
\newblock Deep learning in remote sensing: A comprehensive review and list of resources.
\newblock {\em IEEE geoscience and remote sensing magazine}, 5(4):8--36, 2017.

\bibitem{zhang2022consecutive}
Tong Zhang, Peng Gao, Hao Dong, Yin Zhuang, Guanqun Wang, Wei Zhang, and He~Chen.
\newblock Consecutive pre-training: A knowledge transfer learning strategy with relevant unlabeled data for remote sensing domain.
\newblock {\em Remote Sensing}, 14(22):5675, 2022.

\bibitem{chen2022visualgpt}
Jun Chen, Han Guo, Kai Yi, Boyang Li, and Mohamed Elhoseiny.
\newblock Visualgpt: Data-efficient adaptation of pretrained language models for image captioning.
\newblock In {\em Proceedings of the IEEE/CVF Conference on Computer Vision and Pattern Recognition}, pages 18030--18040, 2022.

\bibitem{li2023blip}
Junnan Li, Dongxu Li, Silvio Savarese, and Steven Hoi.
\newblock Blip-2: Bootstrapping language-image pre-training with frozen image encoders and large language models.
\newblock {\em arXiv preprint arXiv:2301.12597}, 2023.

\bibitem{liu2023visual}
Haotian Liu, Chunyuan Li, Qingyang Wu, and Yong~Jae Lee.
\newblock Visual instruction tuning.
\newblock {\em arXiv preprint arXiv:2304.08485}, 2023.

\bibitem{alayrac2022flamingo}
Jean-Baptiste Alayrac, Jeff Donahue, Pauline Luc, Antoine Miech, Iain Barr, Yana Hasson, Karel Lenc, Arthur Mensch, Katherine Millican, Malcolm Reynolds, et~al.
\newblock Flamingo: a visual language model for few-shot learning.
\newblock {\em Advances in Neural Information Processing Systems}, 35:23716--23736, 2022.

\bibitem{zhao2023mmicl}
Haozhe Zhao, Zefan Cai, Shuzheng Si, Xiaojian Ma, Kaikai An, Liang Chen, Zixuan Liu, Sheng Wang, Wenjuan Han, and Baobao Chang.
\newblock Mmicl: Empowering vision-language model with multi-modal in-context learning.
\newblock {\em arXiv preprint arXiv:2309.07915}, 2023.

\bibitem{zhu2023minigpt}
Deyao Zhu, Jun Chen, Xiaoqian Shen, Xiang Li, and Mohamed Elhoseiny.
\newblock Minigpt-4: Enhancing vision-language understanding with advanced large language models.
\newblock {\em arXiv preprint arXiv:2304.10592}, 2023.

\bibitem{hu2023rsgpt}
Yuan Hu, Jianlong Yuan, Congcong Wen, Xiaonan Lu, and Xiang Li.
\newblock Rsgpt: A remote sensing vision language model and benchmark.
\newblock {\em arXiv preprint arXiv:2307.15266}, 2023.

\bibitem{kuckreja2023geochat}
Kartik Kuckreja, Muhammad~Sohail Danish, Muzammal Naseer, Abhijit Das, Salman Khan, and Fahad~Shahbaz Khan.
\newblock Geochat: Grounded large vision-language model for remote sensing, 2023.

\bibitem{cheng2022nwpu}
Qimin Cheng, Haiyan Huang, Yuan Xu, Yuzhuo Zhou, Huanying Li, and Zhongyuan Wang.
\newblock Nwpu-captions dataset and mlca-net for remote sensing image captioning.
\newblock {\em IEEE Transactions on Geoscience and Remote Sensing}, 60:1--19, 2022.

\bibitem{zhang2023multi}
Meimei Zhang, Fang Chen, and Bin Li.
\newblock Multi-step question-driven visual question answering for remote sensing.
\newblock {\em IEEE Transactions on Geoscience and Remote Sensing}, 2023.

\bibitem{10056343}
Yang Zhan, Zhitong Xiong, and Yuan Yuan.
\newblock Rsvg: Exploring data and models for visual grounding on remote sensing data.
\newblock {\em IEEE Transactions on Geoscience and Remote Sensing}, 61:1--13, 2023.

\bibitem{li2020clrs}
Haifeng Li, Hao Jiang, Xin Gu, Jian Peng, Wenbo Li, Liang Hong, and Chao Tao.
\newblock Clrs: Continual learning benchmark for remote sensing image scene classification.
\newblock {\em Sensors}, 20(4):1226, 2020.

\bibitem{zhou2021nasc}
Zhuang Zhou, Shengyang Li, Wei Wu, Weilong Guo, Xuan Li, Guisong Xia, and Zifei Zhao.
\newblock Nasc-tg2: Natural scene classification with tiangong-2 remotely sensed imagery.
\newblock {\em IEEE Journal of Selected Topics in Applied Earth Observations and Remote Sensing}, 14:3228--3242, 2021.

\bibitem{yu2022mar20}
Wenqi Yu, Gong Cheng, Meijun Wang, Yanqing Yao, Xingxing Xie, XW~Yao, and JW~Han.
\newblock Mar20: A benchmark for military aircraft recognition in remote sensing images.
\newblock {\em National Remote Sensing Bulletin}, 2022.

\bibitem{openai2023gpt4}
OpenAI.
\newblock Gpt-4 technical report, 2023.

\bibitem{touvron2023llama_a}
Hugo Touvron, Thibaut Lavril, Gautier Izacard, Xavier Martinet, Marie-Anne Lachaux, Timoth{\'e}e Lacroix, Baptiste Rozi{\`e}re, Naman Goyal, Eric Hambro, Faisal Azhar, et~al.
\newblock Llama: Open and efficient foundation language models.
\newblock {\em arXiv preprint arXiv:2302.13971}, 2023a.

\bibitem{zhang2023llama}
Renrui Zhang, Jiaming Han, Aojun Zhou, Xiangfei Hu, Shilin Yan, Pan Lu, Hongsheng Li, Peng Gao, and Yu~Qiao.
\newblock Llama-adapter: Efficient fine-tuning of language models with zero-init attention.
\newblock {\em arXiv preprint arXiv:2303.16199}, 2023.

\bibitem{touvron2023llama_b}
Hugo Touvron, Louis Martin, Kevin Stone, Peter Albert, Amjad Almahairi, Yasmine Babaei, Nikolay Bashlykov, Soumya Batra, Prajjwal Bhargava, Shruti Bhosale, et~al.
\newblock Llama 2: Open foundation and fine-tuned chat models.
\newblock {\em arXiv preprint arXiv:2307.09288}, 2023b.

\bibitem{peng2023instruction}
Baolin Peng, Chunyuan Li, Pengcheng He, Michel Galley, and Jianfeng Gao.
\newblock Instruction tuning with gpt-4.
\newblock {\em arXiv preprint arXiv:2304.03277}, 2023.

\bibitem{huang2023language}
Shaohan Huang, Li~Dong, Wenhui Wang, Yaru Hao, Saksham Singhal, Shuming Ma, Tengchao Lv, Lei Cui, Owais~Khan Mohammed, Qiang Liu, et~al.
\newblock Language is not all you need: Aligning perception with language models.
\newblock {\em arXiv preprint arXiv:2302.14045}, 2023.

\bibitem{gao2023llama}
Peng Gao, Jiaming Han, Renrui Zhang, Ziyi Lin, Shijie Geng, Aojun Zhou, Wei Zhang, Pan Lu, Conghui He, Xiangyu Yue, et~al.
\newblock Llama-adapter v2: Parameter-efficient visual instruction model.
\newblock {\em arXiv preprint arXiv:2304.15010}, 2023.

\bibitem{ye2023mplug}
Qinghao Ye, Haiyang Xu, Guohai Xu, Jiabo Ye, Ming Yan, Yiyang Zhou, Junyang Wang, Anwen Hu, Pengcheng Shi, Yaya Shi, et~al.
\newblock mplug-owl: Modularization empowers large language models with multimodality.
\newblock {\em arXiv preprint arXiv:2304.14178}, 2023.

\bibitem{lyu2023macaw}
Chenyang Lyu, Minghao Wu, Longyue Wang, Xinting Huang, Bingshuai Liu, Zefeng Du, Shuming Shi, and Zhaopeng Tu.
\newblock Macaw-llm: Multi-modal language modeling with image, audio, video, and text integration.
\newblock {\em arXiv preprint arXiv:2306.09093}, 2023.

\bibitem{tang2023codi}
Zineng Tang, Ziyi Yang, Mahmoud Khademi, Yang Liu, Chenguang Zhu, and Mohit Bansal.
\newblock Codi-2: In-context, interleaved, and interactive any-to-any generation.
\newblock {\em arXiv preprint arXiv:2311.18775}, 2023.

\bibitem{xu2023pointllm}
Runsen Xu, Xiaolong Wang, Tai Wang, Yilun Chen, Jiangmiao Pang, and Dahua Lin.
\newblock Pointllm: Empowering large language models to understand point clouds.
\newblock {\em arXiv preprint arXiv:2308.16911}, 2023.

\bibitem{guo2023point}
Ziyu Guo, Renrui Zhang, Xiangyang Zhu, Yiwen Tang, Xianzheng Ma, Jiaming Han, Kexin Chen, Peng Gao, Xianzhi Li, Hongsheng Li, et~al.
\newblock Point-bind \& point-llm: Aligning point cloud with multi-modality for 3d understanding, generation, and instruction following.
\newblock {\em arXiv preprint arXiv:2309.00615}, 2023.

\bibitem{hong20233d}
Yining Hong, Haoyu Zhen, Peihao Chen, Shuhong Zheng, Yilun Du, Zhenfang Chen, and Chuang Gan.
\newblock 3d-llm: Injecting the 3d world into large language models.
\newblock {\em arXiv preprint arXiv:2307.12981}, 2023.

\bibitem{yang2023lidar}
Senqiao Yang, Jiaming Liu, Ray Zhang, Mingjie Pan, Zoey Guo, Xiaoqi Li, Zehui Chen, Peng Gao, Yandong Guo, and Shanghang Zhang.
\newblock Lidar-llm: Exploring the potential of large language models for 3d lidar understanding.
\newblock {\em arXiv preprint arXiv:2312.14074}, 2023.

\bibitem{ahn2024autort}
Michael Ahn, Debidatta Dwibedi, Chelsea Finn, Montse~Gonzalez Arenas, Keerthana Gopalakrishnan, Karol Hausman, Brian Ichter, Alex Irpan, Nikhil Joshi, Ryan Julian, et~al.
\newblock Autort: Embodied foundation models for large scale orchestration of robotic agents.
\newblock {\em arXiv preprint arXiv:2401.12963}, 2024.

\bibitem{sun2022ringmo}
Xian Sun, Peijin Wang, Wanxuan Lu, Zicong Zhu, Xiaonan Lu, Qibin He, Junxi Li, Xuee Rong, Zhujun Yang, Hao Chang, et~al.
\newblock Ringmo: A remote sensing foundation model with masked image modeling.
\newblock {\em IEEE Transactions on Geoscience and Remote Sensing}, 2022.

\bibitem{cong2022satmae}
Yezhen Cong, Samar Khanna, Chenlin Meng, Patrick Liu, Erik Rozi, Yutong He, Marshall Burke, David Lobell, and Stefano Ermon.
\newblock Satmae: Pre-training transformers for temporal and multi-spectral satellite imagery.
\newblock {\em Advances in Neural Information Processing Systems}, 35:197--211, 2022.

\bibitem{liu2023remoteclip}
Fan Liu, Delong Chen, Zhangqingyun Guan, Xiaocong Zhou, Jiale Zhu, and Jun Zhou.
\newblock Remoteclip: A vision language foundation model for remote sensing.
\newblock {\em arXiv preprint arXiv:2306.11029}, 2023.

\bibitem{instructblip}
Wenliang Dai, Junnan Li, Dongxu Li, Anthony Meng~Huat Tiong, Junqi Zhao, Weisheng Wang, Boyang Li, Pascale Fung, and Steven Hoi.
\newblock Instructblip: Towards general-purpose vision-language models with instruction tuning, 2023.

\bibitem{lu2017exploring}
Xiaoqiang Lu, Binqiang Wang, Xiangtao Zheng, and Xuelong Li.
\newblock Exploring models and data for remote sensing image caption generation.
\newblock {\em IEEE Transactions on Geoscience and Remote Sensing}, 56(4):2183--2195, 2017.

\bibitem{zheng2021mutual}
Xiangtao Zheng, Binqiang Wang, Xingqian Du, and Xiaoqiang Lu.
\newblock Mutual attention inception network for remote sensing visual question answering.
\newblock {\em IEEE Transactions on Geoscience and Remote Sensing}, 60:1--14, 2021.

\bibitem{lobry2020rsvqa}
Sylvain Lobry, Diego Marcos, Jesse Murray, and Devis Tuia.
\newblock Rsvqa: Visual question answering for remote sensing data.
\newblock {\em IEEE Transactions on Geoscience and Remote Sensing}, 58(12):8555--8566, 2020.

\bibitem{rahnemoonfar2021floodnet}
Maryam Rahnemoonfar, Tashnim Chowdhury, Argho Sarkar, Debvrat Varshney, Masoud Yari, and Robin~Roberson Murphy.
\newblock Floodnet: A high resolution aerial imagery dataset for post flood scene understanding.
\newblock {\em IEEE Access}, 9:89644--89654, 2021.

\bibitem{xia2017aid}
Gui-Song Xia, Jingwen Hu, Fan Hu, Baoguang Shi, Xiang Bai, Yanfei Zhong, Liangpei Zhang, and Xiaoqiang Lu.
\newblock Aid: A benchmark data set for performance evaluation of aerial scene classification.
\newblock {\em IEEE Transactions on Geoscience and Remote Sensing}, 55(7):3965--3981, 2017.

\bibitem{helber2019eurosat}
Patrick Helber, Benjamin Bischke, Andreas Dengel, and Damian Borth.
\newblock Eurosat: A novel dataset and deep learning benchmark for land use and land cover classification.
\newblock {\em IEEE Journal of Selected Topics in Applied Earth Observations and Remote Sensing}, 12(7):2217--2226, 2019.

\bibitem{7891544}
Gong Cheng, Junwei Han, and Xiaoqiang Lu.
\newblock Remote sensing image scene classification: Benchmark and state of the art.
\newblock {\em Proceedings of the IEEE}, 105(10):1865--1883, 2017.

\bibitem{yang2010bag}
Yi~Yang and Shawn Newsam.
\newblock Bag-of-visual-words and spatial extensions for land-use classification.
\newblock In {\em Proceedings of the 18th SIGSPATIAL international conference on advances in geographic information systems}, pages 270--279, 2010.

\bibitem{dai2010satellite}
Dengxin Dai and Wen Yang.
\newblock Satellite image classification via two-layer sparse coding with biased image representation.
\newblock {\em IEEE Geoscience and remote sensing letters}, 8(1):173--176, 2010.

\bibitem{han2014object}
Junwei Han, Dingwen Zhang, Gong Cheng, Lei Guo, and Jinchang Ren.
\newblock Object detection in optical remote sensing images based on weakly supervised learning and high-level feature learning.
\newblock {\em IEEE Transactions on Geoscience and Remote Sensing}, 53(6):3325--3337, 2014.

\bibitem{9560031}
Jian Ding, Nan Xue, Gui-Song Xia, Xiang Bai, Wen Yang, Michael Yang, Serge Belongie, Jiebo Luo, Mihai Datcu, Marcello Pelillo, and Liangpei Zhang.
\newblock Object detection in aerial images: A large-scale benchmark and challenges.
\newblock {\em IEEE Transactions on Pattern Analysis and Machine Intelligence}, pages 1--1, 2021.

\bibitem{sun2022fair1m}
Xian Sun, Peijin Wang, Zhiyuan Yan, Feng Xu, Ruiping Wang, Wenhui Diao, Jin Chen, Jihao Li, Yingchao Feng, Tao Xu, et~al.
\newblock Fair1m: A benchmark dataset for fine-grained object recognition in high-resolution remote sensing imagery.
\newblock {\em ISPRS Journal of Photogrammetry and Remote Sensing}, 184:116--130, 2022.

\bibitem{zhang2019hierarchical}
Yuanlin Zhang, Yuan Yuan, Yachuang Feng, and Xiaoqiang Lu.
\newblock Hierarchical and robust convolutional neural network for very high-resolution remote sensing object detection.
\newblock {\em IEEE Transactions on Geoscience and Remote Sensing}, 57(8):5535--5548, 2019.

\bibitem{li2017rotation}
Ke~Li, Gong Cheng, Shuhui Bu, and Xiong You.
\newblock Rotation-insensitive and context-augmented object detection in remote sensing images.
\newblock {\em IEEE Transactions on Geoscience and Remote Sensing}, 56(4):2337--2348, 2017.

\bibitem{xian2019air}
SUN Xian, WANG Zhirui, SUN Yuanrui, DIAO Wenhui, ZHANG Yue, and FU~Kun.
\newblock Air-sarship-1.0: High-resolution sar ship detection dataset.
\newblock {\em Journal of Radars}, 8(6):852--863, 2019.

\bibitem{wei2020hrsid}
Shunjun Wei, Xiangfeng Zeng, Qizhe Qu, Mou Wang, Hao Su, and Jun Shi.
\newblock Hrsid: A high-resolution sar images dataset for ship detection and instance segmentation.
\newblock {\em Ieee Access}, 8:120234--120254, 2020.

\bibitem{zhang2021sar}
Tianwen Zhang, Xiaoling Zhang, Jianwei Li, Xiaowo Xu, Baoyou Wang, Xu~Zhan, Yanqin Xu, Xiao Ke, Tianjiao Zeng, Hao Su, et~al.
\newblock Sar ship detection dataset (ssdd): Official release and comprehensive data analysis.
\newblock {\em Remote Sensing}, 13(18):3690, 2021.

\bibitem{waqas2019isaid}
Syed Waqas~Zamir, Aditya Arora, Akshita Gupta, Salman Khan, Guolei Sun, Fahad Shahbaz~Khan, Fan Zhu, Ling Shao, Gui-Song Xia, and Xiang Bai.
\newblock isaid: A large-scale dataset for instance segmentation in aerial images.
\newblock In {\em Proceedings of the IEEE Conference on Computer Vision and Pattern Recognition Workshops}, pages 28--37, 2019.

\bibitem{NEURIPS-DATASETS-AND-BENCHMARKS2021_4e732ced}
Junjue Wang, Zhuo Zheng, Ailong Ma, Xiaoyan Lu, and Yanfei Zhong.
\newblock Loveda: A remote sensing land-cover dataset for domain adaptive semantic segmentation.
\newblock In J.~Vanschoren and S.~Yeung, editors, {\em Proceedings of the Neural Information Processing Systems Track on Datasets and Benchmarks}, volume~1. Curran Associates, Inc., 2021.

\bibitem{qu2016deep}
Bo~Qu, Xuelong Li, Dacheng Tao, and Xiaoqiang Lu.
\newblock Deep semantic understanding of high resolution remote sensing image.
\newblock In {\em 2016 International conference on computer, information and telecommunication systems (Cits)}, pages 1--5. IEEE, 2016.

\bibitem{yuan2022exploring}
Zhiqiang Yuan, Wenkai Zhang, Kun Fu, Xuan Li, Chubo Deng, Hongqi Wang, and Xian Sun.
\newblock Exploring a fine-grained multiscale method for cross-modal remote sensing image retrieval.
\newblock {\em arXiv preprint arXiv:2204.09868}, 2022.

\bibitem{dosovitskiy2020image}
Alexey Dosovitskiy, Lucas Beyer, Alexander Kolesnikov, Dirk Weissenborn, Xiaohua Zhai, Thomas Unterthiner, Mostafa Dehghani, Matthias Minderer, Georg Heigold, Sylvain Gelly, et~al.
\newblock An image is worth 16x16 words: Transformers for image recognition at scale.
\newblock {\em arXiv preprint arXiv:2010.11929}, 2020.

\bibitem{o2015introduction}
Keiron O'Shea and Ryan Nash.
\newblock An introduction to convolutional neural networks.
\newblock {\em arXiv preprint arXiv:1511.08458}, 2015.

\bibitem{schuhmann2021laion}
Christoph Schuhmann, Richard Vencu, Romain Beaumont, Robert Kaczmarczyk, Clayton Mullis, Aarush Katta, Theo Coombes, Jenia Jitsev, and Aran Komatsuzaki.
\newblock Laion-400m: Open dataset of clip-filtered 400 million image-text pairs.
\newblock {\em arXiv preprint arXiv:2111.02114}, 2021.

\bibitem{chen2015microsoft}
Xinlei Chen, Hao Fang, Tsung-Yi Lin, Ramakrishna Vedantam, Saurabh Gupta, Piotr Doll{\'a}r, and C~Lawrence Zitnick.
\newblock Microsoft coco captions: Data collection and evaluation server.
\newblock {\em arXiv preprint arXiv:1504.00325}, 2015.

\bibitem{touvron2023llama}
Hugo Touvron, Thibaut Lavril, Gautier Izacard, Xavier Martinet, Marie-Anne Lachaux, Timoth{\'e}e Lacroix, Baptiste Rozi{\`e}re, Naman Goyal, Eric Hambro, Faisal Azhar, et~al.
\newblock Llama: Open and efficient foundation language models.
\newblock {\em arXiv preprint arXiv:2302.13971}, 2023.

\bibitem{zou2015deep}
Qin Zou, Lihao Ni, Tong Zhang, and Qian Wang.
\newblock Deep learning based feature selection for remote sensing scene classification.
\newblock {\em IEEE Geoscience and remote sensing letters}, 12(11):2321--2325, 2015.

\bibitem{di2021public}
Yanghua Di, Zhiguo Jiang, and Haopeng Zhang.
\newblock A public dataset for fine-grained ship classification in optical remote sensing images.
\newblock {\em Remote Sensing}, 13(4):747, 2021.

\bibitem{di2019public}
Yanghua Di, Zhiguo Jiang, Haopeng Zhang, and Gang Meng.
\newblock A public dataset for ship classification in remote sensing images.
\newblock In {\em Image and Signal Processing for Remote Sensing XXV}, volume 11155, pages 515--521. SPIE, 2019.

\bibitem{long2017accurate}
Yang Long, Yiping Gong, Zhifeng Xiao, and Qing Liu.
\newblock Accurate object localization in remote sensing images based on convolutional neural networks.
\newblock {\em IEEE Transactions on Geoscience and Remote Sensing}, 55(5):2486--2498, 2017.

\bibitem{zhu2015orientation}
Haigang Zhu, Xiaogang Chen, Weiqun Dai, Kun Fu, Qixiang Ye, and Jianbin Jiao.
\newblock Orientation robust object detection in aerial images using deep convolutional neural network.
\newblock In {\em 2015 IEEE International Conference on Image Processing (ICIP)}, pages 3735--3739. IEEE, 2015.

\bibitem{zhu2021detection}
Pengfei Zhu, Longyin Wen, Dawei Du, Xiao Bian, Heng Fan, Qinghua Hu, and Haibin Ling.
\newblock Detection and tracking meet drones challenge.
\newblock {\em IEEE Transactions on Pattern Analysis and Machine Intelligence}, 44(11):7380--7399, 2021.

\bibitem{suo2023hit}
Jiashun Suo, Tianyi Wang, Xingzhou Zhang, Haiyang Chen, Wei Zhou, and Weisong Shi.
\newblock Hit-uav: A high-altitude infrared thermal dataset for unmanned aerial vehicle-based object detection.
\newblock {\em Scientific Data}, 10(1):227, 2023.

\bibitem{Sea-shipping}
InfiRay.
\newblock Sea-shipping.
\newblock \url{http://openai.iraytek.com/apply/Sea_shipping.html/}, 2021.

\bibitem{Infrared-security}
InfiRay.
\newblock Infrared-security.
\newblock \url{http://openai.iraytek.com/apply/Infrared_security.html/}, 2021.

\bibitem{InfiRay-Aerial-mancar}
InfiRay.
\newblock Aerial-mancar.
\newblock \url{http://openai.raytrontek.com/apply/Aerial_mancar.html/}, 2021.

\bibitem{Double-light-vehicle}
InfiRay.
\newblock Double-light-vehicle.
\newblock \url{http://openai.raytrontek.com/apply/Double_light_vehicle.html/}, 2021.

\bibitem{oceanic-ship}
Center for Optics~Research and Engineering of~Shandong~University.
\newblock oceanic-ship.
\newblock \url{http://www.gxzx.sdu.edu.cn/info/1133/2174.htm/}, 2020.

\bibitem{oquab2023dinov2}
Maxime Oquab, Timothée Darcet, Theo Moutakanni, Huy~V. Vo, Marc Szafraniec, Vasil Khalidov, Pierre Fernandez, Daniel Haziza, Francisco Massa, Alaaeldin El-Nouby, Russell Howes, Po-Yao Huang, Hu~Xu, Vasu Sharma, Shang-Wen Li, Wojciech Galuba, Mike Rabbat, Mido Assran, Nicolas Ballas, Gabriel Synnaeve, Ishan Misra, Herve Jegou, Julien Mairal, Patrick Labatut, Armand Joulin, and Piotr Bojanowski.
\newblock Dinov2: Learning robust visual features without supervision, 2023.

\bibitem{ilharco_gabriel_2021_5143773}
Gabriel Ilharco, Mitchell Wortsman, Ross Wightman, Cade Gordon, Nicholas Carlini, Rohan Taori, Achal Dave, Vaishaal Shankar, Hongseok Namkoong, John Miller, Hannaneh Hajishirzi, Ali Farhadi, and Ludwig Schmidt.
\newblock Openclip, July 2021.
\newblock If you use this software, please cite it as below.

\bibitem{manas2021seasonal}
Oscar Manas, Alexandre Lacoste, Xavier Gir{\'o}-i Nieto, David Vazquez, and Pau Rodriguez.
\newblock Seasonal contrast: Unsupervised pre-training from uncurated remote sensing data.
\newblock In {\em Proceedings of the IEEE/CVF International Conference on Computer Vision}, pages 9414--9423, 2021.

\bibitem{10184498}
Junjie Wang, Wei Li, Mengmeng Zhang, Ran Tao, and Jocelyn Chanussot.
\newblock Remote-sensing scene classification via multistage self-guided separation network.
\newblock {\em IEEE Transactions on Geoscience and Remote Sensing}, 61:1--12, 2023.

\bibitem{wang2021csds}
Xinyu Wang, Liming Yuan, Haixia Xu, and Xianbin Wen.
\newblock Csds: End-to-end aerial scenes classification with depthwise separable convolution and an attention mechanism.
\newblock {\em IEEE Journal of Selected Topics in Applied Earth Observations and Remote Sensing}, 14:10484--10499, 2021.

\bibitem{wang2022transferring}
Weiquan Wang, Yushi Chen, and Pedram Ghamisi.
\newblock Transferring cnn with adaptive learning for remote sensing scene classification.
\newblock {\em IEEE Transactions on Geoscience and Remote Sensing}, 60:1--18, 2022.

\bibitem{cheng2021perturbation}
Gong Cheng, Xuxiang Sun, Ke~Li, Lei Guo, and Junwei Han.
\newblock Perturbation-seeking generative adversarial networks: A defense framework for remote sensing image scene classification.
\newblock {\em IEEE Transactions on Geoscience and Remote Sensing}, 60:1--11, 2021.

\bibitem{bai2023qwen}
Jinze Bai, Shuai Bai, Shusheng Yang, Shijie Wang, Sinan Tan, Peng Wang, Junyang Lin, Chang Zhou, and Jingren Zhou.
\newblock Qwen-vl: A frontier large vision-language model with versatile abilities.
\newblock {\em arXiv preprint arXiv:2308.12966}, 2023.

\bibitem{lin2023sphinx}
Ziyi Lin, Chris Liu, Renrui Zhang, Peng Gao, Longtian Qiu, Han Xiao, Han Qiu, Chen Lin, Wenqi Shao, Keqin Chen, et~al.
\newblock Sphinx: The joint mixing of weights, tasks, and visual embeddings for multi-modal large language models.
\newblock {\em arXiv preprint arXiv:2311.07575}, 2023.

\bibitem{8633358}
Binqiang Wang, Xiaoqiang Lu, Xiangtao Zheng, and Xuelong Li.
\newblock Semantic descriptions of high-resolution remote sensing images.
\newblock {\em IEEE Geoscience and Remote Sensing Letters}, 16(8):1274--1278, 2019.

\bibitem{zhang2019description}
Xiangrong Zhang, Xin Wang, Xu~Tang, Huiyu Zhou, and Chen Li.
\newblock Description generation for remote sensing images using attribute attention mechanism.
\newblock {\em Remote Sensing}, 11(6):612, 2019.

\bibitem{9866055}
Qimin Cheng, Haiyan Huang, Yuan Xu, Yuzhuo Zhou, Huanying Li, and Zhongyuan Wang.
\newblock Nwpu-captions dataset and mlca-net for remote sensing image captioning.
\newblock {\em IEEE Transactions on Geoscience and Remote Sensing}, 60:1--19, 2022.

\bibitem{marino2019ok}
Kenneth Marino, Mohammad Rastegari, Ali Farhadi, and Roozbeh Mottaghi.
\newblock Ok-vqa: A visual question answering benchmark requiring external knowledge.
\newblock In {\em Proceedings of the IEEE/cvf conference on computer vision and pattern recognition}, pages 3195--3204, 2019.

\bibitem{kafle2016answer}
Kushal Kafle and Christopher Kanan.
\newblock Answer-type prediction for visual question answering.
\newblock In {\em Proceedings of the IEEE conference on computer vision and pattern recognition}, pages 4976--4984, 2016.

\bibitem{chen2023minigpt}
Jun Chen, Deyao Zhu, Xiaoqian Shen, Xiang Li, Zechun Liu, Pengchuan Zhang, Raghuraman Krishnamoorthi, Vikas Chandra, Yunyang Xiong, and Mohamed Elhoseiny.
\newblock Minigpt-v2: large language model as a unified interface for vision-language multi-task learning.
\newblock {\em arXiv preprint arXiv:2310.09478}, 2023.

\bibitem{sadhu2019zero}
Arka Sadhu, Kan Chen, and Ram Nevatia.
\newblock Zero-shot grounding of objects from natural language queries.
\newblock In {\em Proceedings of the IEEE/CVF International Conference on Computer Vision}, pages 4694--4703, 2019.

\bibitem{yang2019fast}
Zhengyuan Yang, Boqing Gong, Liwei Wang, Wenbing Huang, Dong Yu, and Jiebo Luo.
\newblock A fast and accurate one-stage approach to visual grounding.
\newblock In {\em Proceedings of the IEEE/CVF International Conference on Computer Vision}, pages 4683--4693, 2019.

\bibitem{yang2020improving}
Zhengyuan Yang, Tianlang Chen, Liwei Wang, and Jiebo Luo.
\newblock Improving one-stage visual grounding by recursive sub-query construction.
\newblock In {\em Computer Vision--ECCV 2020: 16th European Conference, Glasgow, UK, August 23--28, 2020, Proceedings, Part XIV 16}, pages 387--404. Springer, 2020.

\bibitem{huang2021look}
Binbin Huang, Dongze Lian, Weixin Luo, and Shenghua Gao.
\newblock Look before you leap: Learning landmark features for one-stage visual grounding.
\newblock In {\em Proceedings of the IEEE/CVF Conference on Computer Vision and Pattern Recognition}, pages 16888--16897, 2021.

\bibitem{deng2021transvg}
Jiajun Deng, Zhengyuan Yang, Tianlang Chen, Wengang Zhou, and Houqiang Li.
\newblock Transvg: End-to-end visual grounding with transformers.
\newblock In {\em Proceedings of the IEEE/CVF International Conference on Computer Vision}, pages 1769--1779, 2021.

\bibitem{yang2022improving}
Li~Yang, Yan Xu, Chunfeng Yuan, Wei Liu, Bing Li, and Weiming Hu.
\newblock Improving visual grounding with visual-linguistic verification and iterative reasoning.
\newblock In {\em Proceedings of the IEEE/CVF Conference on Computer Vision and Pattern Recognition}, pages 9499--9508, 2022.

\bibitem{liu2023grounding}
Shilong Liu, Zhaoyang Zeng, Tianhe Ren, Feng Li, Hao Zhang, Jie Yang, Chunyuan Li, Jianwei Yang, Hang Su, Jun Zhu, et~al.
\newblock Grounding dino: Marrying dino with grounded pre-training for open-set object detection.
\newblock {\em arXiv preprint arXiv:2303.05499}, 2023.

\bibitem{zhao2024open}
Xiangyu Zhao, Yicheng Chen, Shilin Xu, Xiangtai Li, Xinjiang Wang, Yining Li, and Haian Huang.
\newblock An open and comprehensive pipeline for unified object grounding and detection.
\newblock {\em arXiv preprint arXiv:2401.02361}, 2024.

\bibitem{wei2023lenna}
Fei Wei, Xinyu Zhang, Ailing Zhang, Bo~Zhang, and Xiangxiang Chu.
\newblock Lenna: Language enhanced reasoning detection assistant, 2023.

\bibitem{han2021align}
Jiaming Han, Jian Ding, Jie Li, and Gui-Song Xia.
\newblock Align deep features for oriented object detection.
\newblock {\em IEEE Transactions on Geoscience and Remote Sensing}, 2021.

\bibitem{guo2021beyond}
Zonghao Guo, Chang Liu, Xiaosong Zhang, Jianbin Jiao, Xiangyang Ji, and Qixiang Ye.
\newblock Beyond bounding-box: Convex-hull feature adaptation for oriented and densely packed object detection.
\newblock In {\em Proceedings of the IEEE/CVF conference on Computer Vision and Pattern Recognition}, pages 8792--8801, 2021.

\bibitem{li2022oriented}
Wentong Li, Yijie Chen, Kaixuan Hu, and Jianke Zhu.
\newblock Oriented reppoints for aerial object detection.
\newblock In {\em Proceedings of the IEEE/CVF conference on computer vision and pattern recognition}, pages 1829--1838, 2022.

\bibitem{xie2021oriented}
Xingxing Xie, Gong Cheng, Jiabao Wang, Xiwen Yao, and Junwei Han.
\newblock Oriented r-cnn for object detection.
\newblock In {\em Proceedings of the IEEE/CVF international conference on computer vision}, pages 3520--3529, 2021.

\bibitem{hou2022shape}
Liping Hou, Ke~Lu, Jian Xue, and Yuqiu Li.
\newblock Shape-adaptive selection and measurement for oriented object detection.
\newblock In {\em Proceedings of the AAAI Conference on Artificial Intelligence}, 2022.

\bibitem{yang2021r3det}
Xue Yang, Junchi Yan, Ziming Feng, and Tao He.
\newblock R3det: Refined single-stage detector with feature refinement for rotating object.
\newblock In {\em Proceedings of the AAAI Conference on Artificial Intelligence}, volume~35, pages 3163--3171, 2021.

\end{thebibliography}

\end{document}